\documentclass[10pt,twocolumn,letterpaper]{article}
\usepackage[pagenumbers]{cvpr}
\usepackage{booktabs}
\usepackage[table]{xcolor}
\usepackage[accsupp]{axessibility}

\usepackage{amsmath}
\usepackage{amssymb}
\usepackage{array}
\usepackage{bm}
\usepackage{booktabs}
\usepackage{caption}
\usepackage{fancyvrb}
\usepackage[flushmargin]{footmisc}
\usepackage{gensymb}
\usepackage{graphicx}
\usepackage{lipsum}
\usepackage{multirow}
\usepackage{pifont}
\usepackage{tcolorbox}
\usepackage{wrapfig}
\usepackage{xcolor}
\usepackage{xspace}

\definecolor{cvprblue}{rgb}{0.21,0.49,0.74}
\usepackage[pagebackref,breaklinks,colorlinks,allcolors=cvprblue]{hyperref}

\tcbuselibrary{listings, breakable}

\newcommand{\method}{VGGT-$\Omega$\xspace}

\newcommand{\methodit}{\textit{VGGT-}\textit{$\varOmega$}\xspace}

\definecolor{promptcolor}{RGB}{20, 120, 20}
\definecolor{linegreen}{RGB}{20, 140, 50}

\newcommand{\duster}{DUSt3R\xspace}
\newcommand{\master}{MASt3R\xspace}

\newcommand{\bt}{\boldsymbol{t}}
\newcommand{\bq}{\boldsymbol{q}}
\newcommand{\bz}{\boldsymbol{z}}

\newcommand{\bg}{\boldsymbol{g}}
\newcommand{\bbf}{\boldsymbol{f}}
\newcommand{\R}{\mathbb{R}}

\makeatletter
\renewcommand{\paragraph}{%
    \@startsection{paragraph}{4}%
    {\z@}{-0.45em}{-0.45em}%
    {\normalfont\normalsize\bfseries}%
}
\makeatother

\def\paperID{11696}
\def\confName{CVPR}
\def\confYear{2026}

\title{\method}

\author{
Jianyuan Wang$^{1,2}$
\vspace{0.02cm}%
\and
Minghao Chen$^{1}$
\and
Shangzhan Zhang$^{1}$
\and
Nikita Karaev$^{1}$
\and
Johannes Sch\"onberger$^{2}$%
\vspace{0.02cm}%
\and
Patrick Labatut$^{2}$
\and
Piotr Bojanowski$^{2}$
\and
David Novotny
\and
Andrea Vedaldi$^{1,2}$
\and
Christian Rupprecht$^{1}$
\and
\\
$^{1}$Visual Geometry Group, University of Oxford \hspace{5em}
$^{2}$Meta AI
}

\begin{document}
\maketitle

\begin{abstract}
Recent feed-forward reconstruction models, such as VGGT, have proven competitive with traditional optimization-based reconstructors while also providing geometry-aware features useful for other tasks.
Here, we show that the quality of these models scales predictably with model and data size.
We do so by introducing \methodit, which substantially improves reconstruction accuracy, efficiency, and capabilities for both static and dynamic scenes.
To enable training this model at an unprecedented scale, we introduce architectural changes that improve training efficiency, a high-quality data annotation pipeline that supports dynamic scenes, and a self-supervised learning protocol.
We simplify VGGT's architecture by using a single dense prediction head with multi-task supervision and removing the expensive high-resolution convolutional layers.
We also use registers to aggregate scene information into a compact representation and introduce register attention, which restricts inter-frame information exchange to these registers, in part replacing global attention.
In this way, during training, \methodit uses only $\sim$30\% of the GPU memory of its predecessor, which allows us to train \methodit with 15$\times$ more supervised data than prior work and to leverage vast amounts of unlabeled video data.
\methodit achieves strong results for reconstruction of static and dynamic scenes across multiple benchmarks, \eg, improving over the previous best camera estimation accuracy on Sintel by 77\%.
We also show that the learned registers can improve vision-language-action models and support alignment with language, suggesting that reconstruction can be a powerful and scalable proxy task for spatial understanding.
\par\noindent Project page: \url{http://vggt-omega.github.io/}
\end{abstract}

\vspace{-8pt}
\section{Introduction}%
\label{sec:intro}

\begin{figure}
\centering
\includegraphics[width=0.95\linewidth]{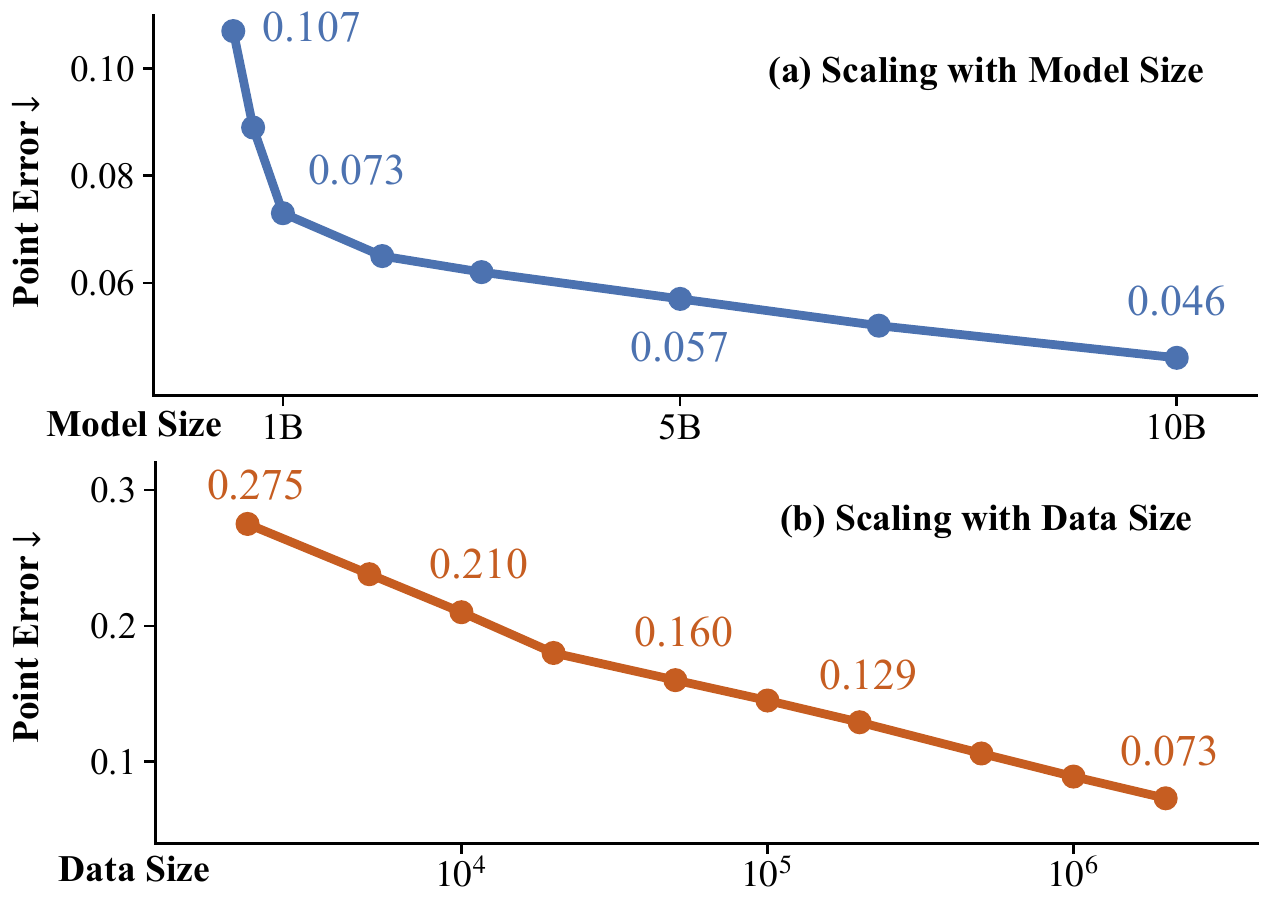}
\caption{\textbf{Performance Gains from Data and Model Parameter Scaling.}
As model size increases from 0.2B to 10B parameters and data scale grows from 2K to 2M sequences, performance improves consistently, as measured by 3D point error (lower is better; note the different axis scales).
All models are trained on approximately the same number of tokens and evaluated by averaging over six datasets, with details provided in \cref{sec:benchmarking}.
}%
\label{fig:scale}
\vspace{-10pt}
\end{figure}

Recent work~\cite{wang24dust3r:, leroy24grounding, wang25vggt, wang25p3, lin25depth} has shown that feed-forward reconstruction models can, in many cases, match and even surpass traditional structure-from-motion (SfM) pipelines~\cite{hartley00multiple,snavely06photo,schonberger16structure-from-motion}.
Furthermore, the tokens learned by such models have been used as effective geometry-aware representations in many other tasks~\cite{li25spatial, abouzeid25geoaware-vla:, zeng25janusvln:,qian26xembodied:, zhang26spatialstack:, zhang26make, zheng25learning,bonnen26human-level,wang26let-geometry, huang26gen3r:, jang26repurposing, yang26cambrian-s:, szymanowicz26lagernvs,cao26vggt-det:, gao26vggt-segmentor:, zhao25spacemind:, yang25dense, vuong25improving, rao26augvla-3d:, zhou26generalizing, xu26vggt-mpr:, wu26mvggt:, lee253d-aware}.
This indicates that reconstruction can serve as a proxy task for learning representations useful for spatial understanding in general, with a foundational value.
However, compared to foundation models where the role of scale is well understood~\cite{simeoni25dinov3, kaplan20scaling, zhai22scaling}, this is less explored in 3D computer vision.
In this paper, we therefore ask whether feed-forward reconstruction models can be scaled up, and what benefits such scaling brings.
To answer this question, we introduce \method, scaling feed-forward reconstruction to significantly larger data and, optionally, model size than prior work.

Compared to VGGT~\cite{wang25vggt}, the new model introduces a number of architectural improvements, beginning with how it uses registers.
Recent works~\cite{darcet24vision, jiang25vision, marouani26revisiting} noted that vision transformers (ViTs) spontaneously use a small number of image tokens to carry global information, and introduced learnable registers to do so more directly and efficiently.
While VGGT already has per-frame registers, \method further introduces \emph{register attention}: in a subset of the global attention layers, information exchange among frames is restricted to the registers.
The updated registers then interact with other tokens locally within frame attention layers, thereby forming a bottleneck for aggregating and redistributing multi-frame information.
This design encourages the registers to aggregate information about the scene as a whole, and we also call them `scene' tokens.

There are two benefits to this design.
First, while in other architectures registers are often treated as auxiliary and discarded at inference time~\cite{darcet24vision}, we instead show that they carry useful global information.
In particular, although without explicit supervision, they provide useful features for vision-language-action (VLA) models and language alignment.
Second, register attention also improves efficiency.
Global attention is the main computational bottleneck in VGGT, but its attention maps are very sparse~\cite{shen26fastvggt:, wang25faster}.
We find that register attention, by aggregating global information, can also serve as an efficient substitute for full global attention.
Specifically, replacing 25\% of global attention layers with register attention incurs no measurable performance drop, while saving around 23\% FLOPs and 16\% memory in the backbone during training\footnote{Replacing \textit{all} global attention layers with register attention reduces FLOPs to just 6\% of the original, but leads to a considerable performance drop.}.

Registers aside, we also note that high-resolution convolutional layers in dense prediction heads (\eg, DPT) consume a disproportionate amount of GPU memory for storing forward activations, despite accounting for only a small fraction of the model's parameters.
Techniques like FSDP or gradient checkpointing cannot eliminate the cost of storing these activations.
Instead, our second change is to replace the most memory-intensive convolutional layers in the dense predictors with a single MLP followed by a pixel shuffle operator.
This uses little memory without performance degradation, both quantitatively and qualitatively.

Finally, in VGGT, we showed that multi-task training, where depth maps, point maps, and tracking features are supervised directly, is beneficial.
Here, we find that additional dense \textit{heads} are unnecessary to achieve these benefits.
Our third change is to still use multi-task \textit{losses}, but to retain only a single dense head for depth prediction and a single sparse head for camera prediction.

These three changes save 70\% of GPU memory during training and modestly improve inference speed.

In addition to efficiency, we find that the amount, diversity, and quality of the training data are critical for scaling.
In particular, handling \emph{dynamic} content is essential as it unlocks orders of magnitude more Internet-like videos for training.
Therefore, we develop a high-quality data annotation pipeline that can produce annotations for both rigid and dynamic videos at scale.
The pipeline integrates VLM-based pre-filtering, VGGT, COLMAP, modern image-matching models, and supervised geometric post-filtering.
Applied to around $40$ million internal Internet-style videos, the filtering pipeline retains $0.8$ million sequences with accurate annotations, roughly one-third of which contain dynamic content.
Combined with existing datasets (both real and synthetic), this yields a total of $4$M diverse scenes/sequences with accurate reconstruction annotations, more than $15\times$ as many as VGGT\@.

To further improve generalization, we introduce a self-supervised learning protocol inspired by DINO and related momentum teacher-student methods~\cite{caron21emerging, oquab24dinov2:,simeoni25dinov3, tarvainen17mean, he20momentum}.
We maintain teacher and student models initialized from a supervised \method checkpoint.
Both models process the same input sequences under different augmentations and frame permutations.
The student is trained to match the teacher's predictions and feature distribution (after aligning the frame order), while the teacher is updated via an exponential moving average of the student.
We use this protocol to train on $18$ million unlabeled videos.

These improvements allow us to investigate the scaling properties of feed-forward reconstruction models.
As illustrated in \cref{fig:scale}, we observe a consistent power-law-like improvement in reconstruction accuracy (measured by point error) as we increase the model capacity from 0.2B to 10B parameters and expand the training data from a few thousand to two million different sequences.

Overall, \method delivers a new level of feed-forward reconstruction performance, achieving state-of-the-art results in three static and three dynamic benchmarks by a wide margin.
In particular, it substantially outperforms post-optimization methods such as MegaSaM and recent feed-forward methods such as Depth Anything 3~\cite{lin25depth}.
On Sintel, \method attains AUC@$3^\circ$ of $40.0$ vs.\ $22.5$ (by $77\%$) and AUC@$30^\circ$ of $79.1$ vs.\ $58.3$ (by $35\%$) for camera estimation, as well as $\delta_{1.25}$ of $93.5$ vs.\ $74.1$ (by $26\%$) for depth estimation, while being $50\times$ faster than MegaSaM\@.
Finally, we show that the learned registers can be reused beyond reconstruction, improving VLA models and supporting alignment with language.

\begin{figure*}
\centering
\includegraphics[width=0.95\textwidth]{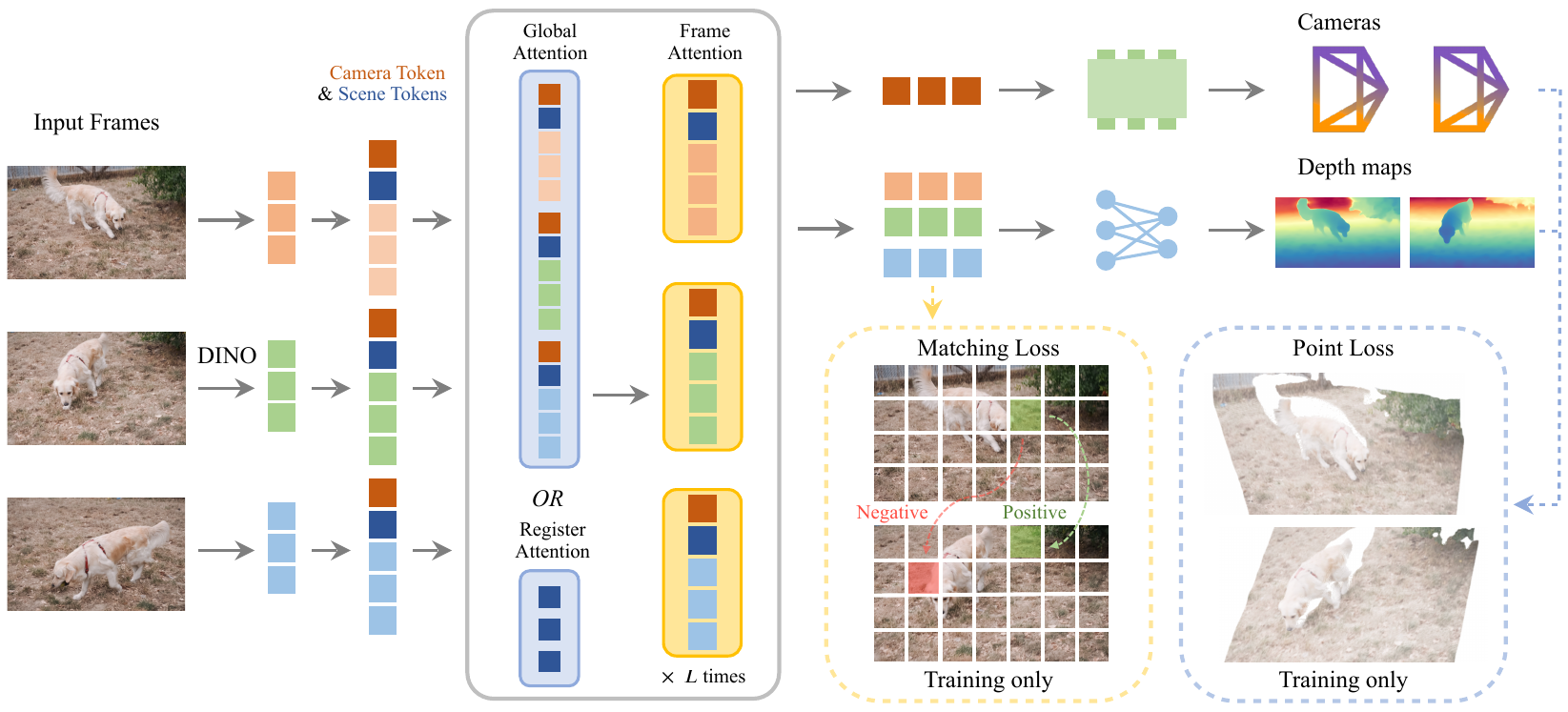}
\vspace{-2pt}
\caption{\textbf{Architecture Overview.} \method appends camera and scene tokens (registers) to image tokens, and then alternates between global attention (or register attention) and frame attention layers.
We replace the redundant dense heads of VGGT with training-only losses.
}%
\label{fig:architecture}
\vspace{-10pt}
\end{figure*}

\section{Related work}%
\label{sec:related}

\paragraph{3D Reconstruction.}

There is a long and rich history of research on 3D reconstruction, beginning with seminal works that established the theory of multi-view geometry~\cite{faugeras90motion, oliensis00a-critique, hartley04multiple, ozyesil17a-survey}.
Follow-up work led to major practical advances, including robust SfM systems such as COLMAP~\cite{schonberger16structure-from-motion} and other pipelines~\cite{snavely06photo, agarwal11building, frahm10building, liu24robust, dellaert12factor}.
In this paper, we focus on \textit{feed-forward reconstruction models}, \ie, neural networks that infer scene geometry and camera poses directly from one or more images.
While recent SfM pipelines increasingly include learnable components such as keypoint detectors~\cite{yi16lift:,daniel18superpoint:,dusmanu19d2-net:,tyszkiewicz20disk:} and feature matchers~\cite{sarlin20superglue:,chen21learning,shi22clustergnn:,lindenberger23lightglue:}, our work is most closely related to end-to-end differentiable SfM frameworks that learn geometry estimation directly~\cite{zhou17unsupervised, ummenhofer17demon:, tang19ba-net:, wei20deepsfm:, wang21deep, teed20deepv2d:, teed21droid-slam:, brachmann24scene, wang23posediffusion:, wang24vggsfm:}.
While these works demonstrate that end-to-end learning is possible in SfM, they still \emph{combine} elements of classical SfM pipelines.
\duster~\cite{wang24dust3r:} and its follow-up \master~\cite{duisterhof25mast3r-sfm:} estimated both scene geometry and camera parameters (extrinsics and intrinsics) directly from images.
However, similar to stereo approaches~\cite{niemeyer20differentiable, fu22geo-neus:, wei21nerfingmvs:, yariv20multiview, yao18mvsnet:, gu20cascade, ma22multiview, peng22rethinking, zhang23geomvsnet:}, the neural networks in \duster and \master operate only on image pairs and still require post-optimization to process additional views.
A key improvement came from methods that process multiple images jointly, removing the need for optimization across views, including~\cite{wang24spann3r:, wang25continuous, tang25mv-dust3r:, wu25point3r:, zhang25flare:, yang25fast3r:, elflein25light3r-sfm:}.
Among these, VGGT~\cite{wang25vggt} is a representative approach that first surpassed post-optimization methods (\eg, using bundle adjustment) while relying solely on feed-forward inference, prompting many follow-up works~\cite{wang25p3, keetha26mapanything:, xiao25spatialtrackerv2:, chen25ttt3r:, deng25vggt-long:, lan26stream3r:, maggio25vggt-slam:, li25wint3r:, huang25longsplat:, szymanowicz26lagernvs, yuan25test3r:, deng26sail-recon:, jiang25anysplat:, cabon25must3r:, liu25plana3r:, li25rig3r:, zang26robust, chen26hd-vggt:, liu26streamcachevggt:, zhuo26streaming, jung26more:, mahdi25evict3r:, wang26lidar-vggt:, qi25fupad:, mazur264d-primitive-mache:}.
Several works improve the computational scalability of VGGT-style models through token merging, sparse attention, or descriptor-based aggregation, making many-view feed-forward reconstruction more efficient~\cite{shen26fastvggt:, wang25faster, mahdi25evict3r:}.
Other works address the quadratic cost of global self-attention by using stateful, bounded scene representations with a linear update cost.
Examples include methods that use test-time-training layers~\cite{chen26ttt3r:, jin26zipmap:, zhang26loger:}.
Others~\cite{zhuo26streaming, lan26stream3r:, chen26geometric, taher25kv-tracker:} process frames causally while maintaining a persistent geometric state by caching tokens, key-value pairs, local windows, or maps.
Some have used VGGT to reconstruct scenes locally, fusing results by means of alignment, localization, or SLAM algorithms~\cite{deng25vggt-long:, maggio25vggt-slam:, deng26sail-recon:, wang25amb3r:}.

Other works have extended VGGT-like models to take additional geometric information as input (\eg, cameras~\cite{keetha26mapanything:}), add sensors (\eg, LiDAR~\cite{wang26lidar-vggt:}), or model camera rigs~\cite{li25rig3r:}.
Other extensions predict normals~\cite{fang26dens3r:} or use mixture-of-experts designs~\cite{gao25more:, wang26moe3d:}.
PI3~\cite{wang25p3} removes the reliance on a fixed reference view, while Depth Anything 3 (DA3)~\cite{lin25depth} adopts vanilla DINO as its backbone, with both supporting dynamic scenes.
Some studies have investigated what these models learn, probing for correspondences, epipolar structure, and connections to 3D shape perception in humans~\cite{bratulic25on-geometric, bonnen26human-level}.
SelfEvo~\cite{huang26self-improving} explored a self-distillation scheme using spatiotemporal context asymmetry. 
Beyond reconstruction itself, the features learned by feed-forward reconstruction models have also been applied as geometry-aware representations for other tasks.
This includes video generation and novel view synthesis~\cite{jiang25anysplat:, huang26gen3r:, jang26repurposing, liu26camgeo:, szymanowicz26lagernvs}, vision-language models~\cite{zeng25janusvln:, qian26xembodied:, zhang26spatialstack:, zhang26make, zheng25learning, wang26let-geometry, yang26cambrian-s:, zhao25spacemind:, wu26mvggt:, lee253d-aware}, vision-language-action models~\cite{li25spatial, abouzeid25geoaware-vla:, vuong25improving, rao26augvla-3d:}, and perception tasks such as detection, segmentation, matching, occupancy prediction, and place recognition~\cite{cao26vggt-det:, gao26vggt-segmentor:, yang25dense, zhou26generalizing, xu26vggt-mpr:, deng25unipr-3d:}.

Monocular dynamic 3D reconstruction, or 4D reconstruction, aims to recover scene geometry that changes over time.
This line of research also has a long history, with early work by Bregler et al.~\cite{bregler00recovering} and Torresani et al.~\cite{torresani08nonrigid}.
Among recent contributions, MegaSaM~\cite{li25megasam:} has been particularly influential, combining feed-forward depth prediction with optimization-based non-rigid reconstruction.
ViPE~\cite{huang25vipe:} further builds on this direction.
Several works have explored feed-forward 4D reconstruction with reduced optimization.
MonST3R~\cite{zhang24monst3r:} and D$^2$USt3R~\cite{han25d2ust3r:} extend \duster to handle dynamic 3D content.
Align3R~\cite{lu25align3r:} builds on \duster to infer cameras and align monocular depth predictions over time, though it still relies on optimization beyond two views.
CUT3R~\cite{wang25continuous} and Point3R~\cite{wu25point3r:} support incremental reconstruction alongside dynamic scenes.
Geo4D~\cite{jiang25geo4d} fine-tunes a video generator to recover 4D geometry.
PI3~\cite{wang25p3} and DA3~\cite{lin25depth} adopt VGGT-style models and train them with dynamic data, while PAGE-4D~\cite{zhou25page-4d:} adapts VGGT through a module that separates static and moving regions.
Human3R focuses on human-scene reconstruction~\cite{chen25human3r:}.
SpatialTracker~\cite{xiao25spatialtrackerv2:}, St4RTrack~\cite{feng25st4rtrack:}, DPM~\cite{sucar25dynamic}, V-DPM~\cite{sucar26v-dpm}, Uni4D~\cite{yao25uni4d:}, Any4D~\cite{karhade2025any4d}, and related methods unify dynamic reconstruction and point tracking.
D4RT~\cite{zhang25efficiently} introduces a unified feed-forward transformer that reconstructs dynamic scenes by querying point-level 4D scene information.

\paragraph{Registers in Vision Transformers (ViTs).}

Recent works~\cite{darcet24vision, marouani26revisiting} have found that a small number of tokens in ViTs encode not only local patch information but also global information.
These outliers have high norms and disrupt the spatial coherence of the patch features.
This, in turn, has motivated the use of register tokens, which are separate from image tokens and help preserve coherence among patch representations. 
Further studies have investigated the causes of this phenomenon and the interactions between register and image tokens~\cite{jiang25vision, shi26vision, lappe25register, marouani26revisiting}.
Here, we add register tokens to each input image frame and use them to aggregate and exchange global information across images.
In this way, our registers carry information about the sequence as a whole.
Rather than discarding them as is often done in prior work, we show that they are useful in downstream applications.

\newcommand{\zscene}{\bz^{\text{scene}}}
\newcommand{\zcamera}{\bz^{\text{cam}}}

\section{Method}%
\label{sec:method}

\method builds on the original VGGT, improving it in several ways.
We begin by describing the new architecture (\cref{sec:architecture}) and training pipeline (\cref{sec:training-losses}) of \method.
Then, we introduce a self-supervised training protocol to exploit unlabeled data (\cref{sec:self-supervised-training}) and a new data pipeline for robustly annotating millions of videos (\cref{sec:data}).

\subsection{A New Scalable Architecture}%
\label{sec:architecture}

\method, illustrated in \cref{fig:architecture}, is a feed-forward transformer $f$ that maps $N$ input images
$
I_1,\dots,I_N \in \mathbb{R}^{3\times H \times W}
$
to corresponding cameras and depth maps:
$$
((\bg_1, D_1),\dots,(\bg_N,D_N))
=
f(I_1,\dots,I_N),
$$
where
$D_i \in \mathbb{R}^{H \times W}$
is the depth map of image $I_i$ and
$\bg_i = (\bq_i, \bt_i, \bbf_i) \in \mathbb{R}^{9}$
is the concatenation of
the {rotation} quaternion $\bq_i \in \R^4$,
the {translation} vector $\bt_i \in \R^3$,
and the {field of view} $\bbf_i \in \R^2$
describing the corresponding camera.
As is commonly done~\cite{schonberger16structure-from-motion,wang24vggsfm:,wang25vggt}, we assume that the principal point is at the center of the image.
The problem formulation is thus similar to VGGT~\cite{wang25vggt}, except that the model \emph{does not} predict point maps or tracking features directly (although it still supervises them, as discussed in \cref{sec:training-losses}).
The network $f$ encodes each image into tokens (\cref{sec:feature-extraction-and-tokenization}), aggregates features across views with alternating-attention (\cref{sec:global-attention}), and maps the tokens to the final predictions using lightweight heads (\cref{sec:heads}).

\subsubsection{Feature Extraction and Tokenization}%
\label{sec:feature-extraction-and-tokenization}

We tokenize each image $I_i$ with a DINOv3-initialized vision transformer~\cite{simeoni25dinov3}, obtaining
$
\bz_i^F = \operatorname{DINO}(I_i) \in \mathbb{R}^{H'W'\times C}
$,
where $H'=H/r$ and $W'=W/r$ for patch size $r$.
For each image $I_i$, we also append one \emph{camera token} $\zcamera_i\in\mathbb{R}^{1\times C}$ and sixteen \emph{registers (scene tokens)} $\zscene_i\in\mathbb{R}^{16\times C}$.
The camera token is used to predict the camera parameters, and the registers aggregate information about the scene.
As in~\cite{wang25vggt}, these tokens can take one of two learnable parameters, one if image $I_i$ is the \emph{reference image} and the other otherwise.
These are then concatenated to form the set of tokens
$
\bz = (\bz_1,\dots,\bz_N) \in \mathbb{R}^{N \times (H'W'+17)\times C}
$
where $\bz_i = (\bz_i^F, \zcamera_i, \zscene_i)$ are the tokens of image $I_i$.

\subsubsection{Register Attention}%
\label{sec:global-attention}

\begin{figure}
\centering
\includegraphics[width=0.98\linewidth]{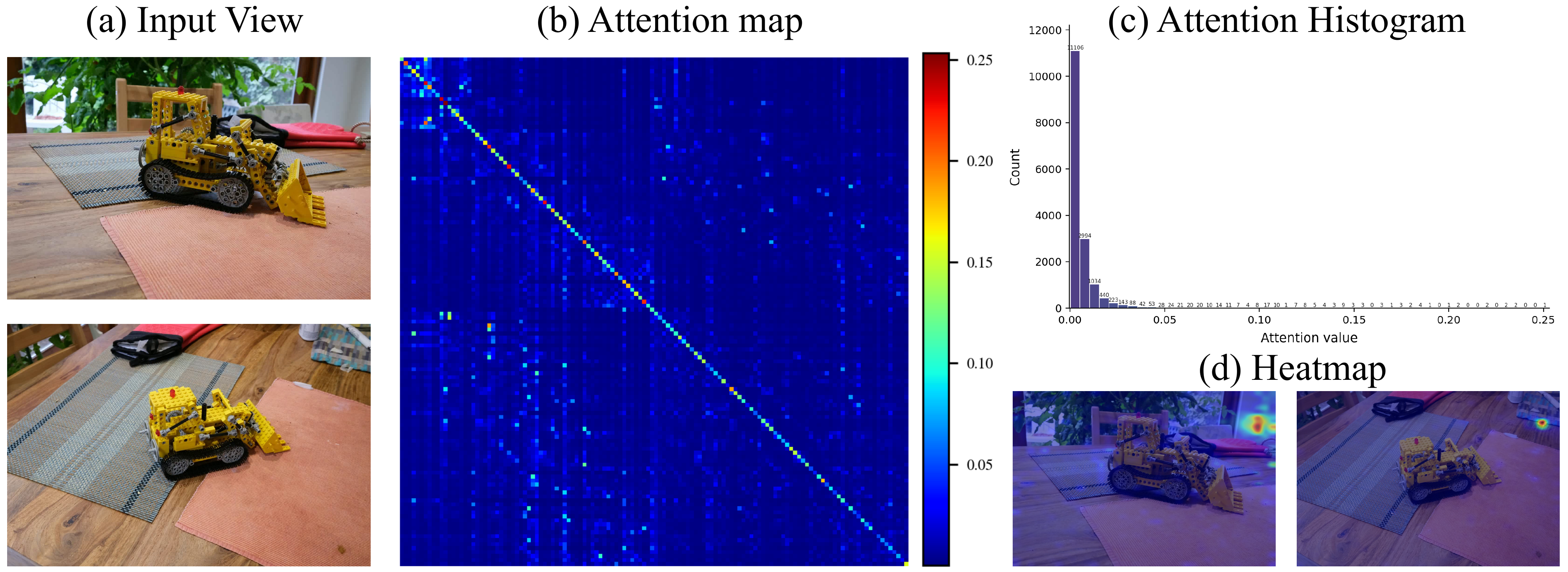}
\caption{\textbf{Visualization of Global Attention in VGGT}. As shown in (b) the global attention matrix, (c) its value distribution, and (d) spatial heatmaps, the attention in layer $13$ is quite sparse.}%
\label{fig:atten_visual}
\vspace{-8pt}
\end{figure}

\begin{figure*}
\centering
\includegraphics[width=\textwidth]{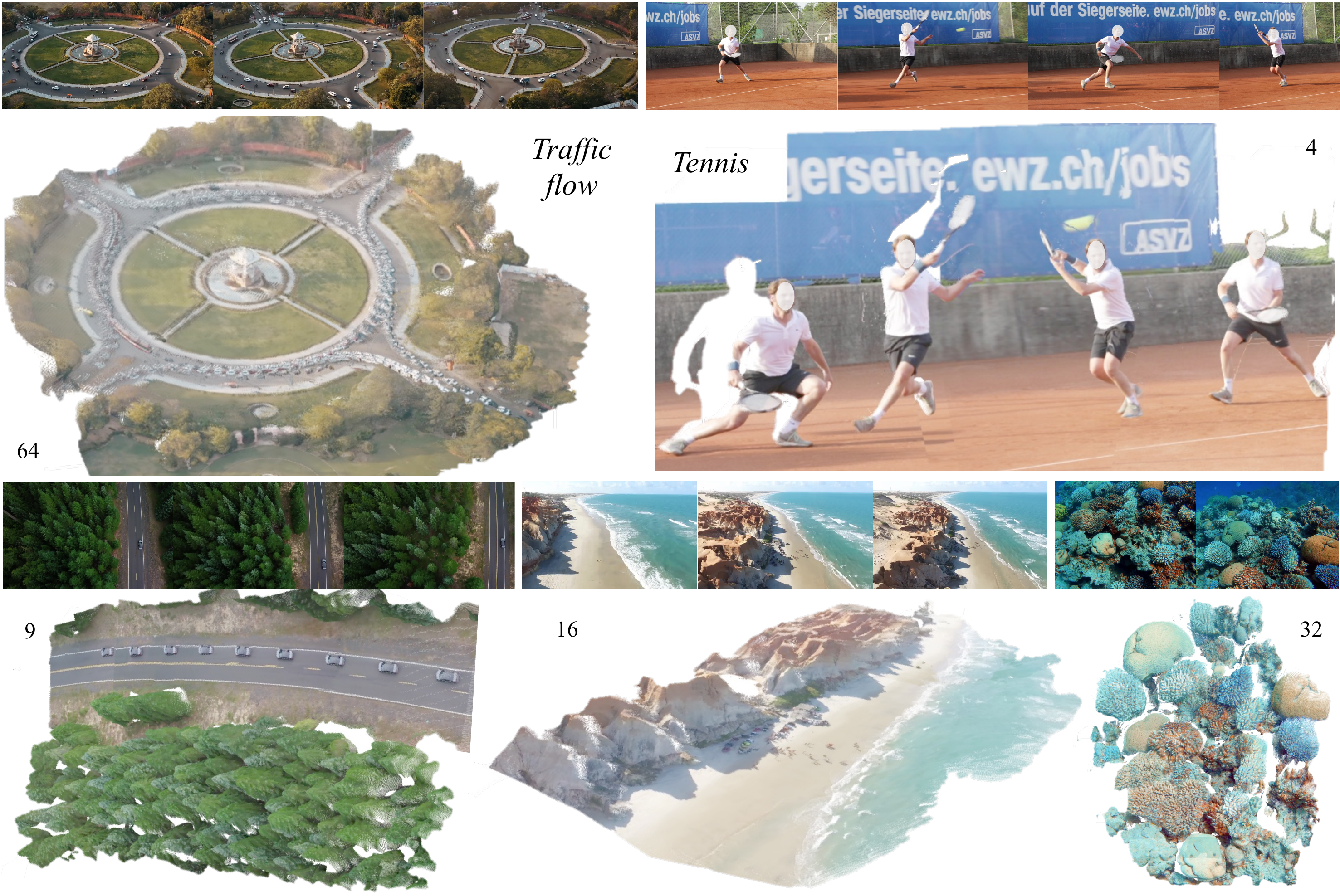}
\caption{\textbf{Qualitative Results.} 
\method handles both static and dynamic content, as evidenced by the overlaid traffic flow and the tennis player's trajectory.
It also generalizes to hard scenes, \eg, underwater coral reefs.
Each example uses $64$, $4$, $9$, $16$, and $32$ input frames.
}%
\label{fig:qual}
\end{figure*}

Recall that VGGT uses \emph{alternating-attention}~\cite{wang25vggt}, interleaving between frame-wise self-attention within each image and global self-attention across all images, which are, by definition, permutation equivariant.
Frame-wise attention eliminates the need for frame-index embeddings, which would otherwise restrict the model's ability to generalize to a flexible number of input frames.
Therefore, none of the tokens has an explicit encoding of the corresponding image identity (except for indicating whether a frame is the reference).
{Global} attention is the standard attention layer applied to all tokens $\bz$, which we denote
$
\bz' = \operatorname{attn}(\bz)
$.
{Frame-wise} attention is similar, but applied independently to $\bz_i$ for each image $I_i$,
which we denote
$
\bz' =
\operatorname{attn}_f(\bz) =
(
    \operatorname{attn}(\bz_1), \dots,
    \operatorname{attn}(\bz_N)
)
$.
Global attention is where different frames interact, and thus where multi-frame scene information is formed.
At the same time, global attention is computationally expensive, as it attends to all tokens from all frames, with a cost that is quadratic in the total number of tokens.
Moreover, we observe that global attention maps are typically sparse, as shown in \cref{fig:atten_visual}, which may suggest that a small number of tokens are enough to exchange the corresponding information.
This is consistent with recent findings~\cite{shen26fastvggt:, wang25faster}.
We therefore replace 25\% of the global attention layers with \emph{register attention}, in which self-attention is restricted to the registers of all frames.
Formally, register attention updates only the registers:
$
\bz' = \operatorname{attn}_\text{scene}(\bz)
$
where
$
({\zscene_1}',\dots,{\zscene_N}') = \operatorname{attn}( \zscene_1,\dots,\zscene_N )
$,
\ie, only the registers participate in self-attention across frames in these blocks.
The updated registers then interact with each frame's image tokens in subsequent frame-wise attention blocks, redistributing the aggregated scene information back to the image tokens.
This encourages the final registers to carry global scene information, while also reducing the cost of global attention.

\subsubsection{Decoding}%
\label{sec:heads}

The final set of tokens $\bz' = (\bz_1',\dots,\bz_N')$ produced by the attention layers is decoded into depth maps and cameras.

\paragraph{Depth.}

In VGGT, all dense decoders are implemented with Dense Prediction Transformer (DPT) layers~\cite{ranftl21vision}.
However, the final convolutional blocks in these DPT heads maintain several high-resolution feature maps, which are memory-intensive.
To reduce this cost, we replace the blocks operating above $1/4$ of the input resolution with a lightweight upsampling head via a single MLP followed by a pixel-shuffle operator.
The MLP outputs $2u^{2}$ channels ($u=4$ in our implementation), and the pixel-shuffle operator rearranges them from $(H' \times W', 2u^{2})$ to $(uH') \times (uW') \times 2$, where the two output channels correspond to depth and confidence.

We also explored a fully convolution-free decoder that maps tokens to dense predictions using MLPs only.
While this works well on benchmarks, qualitatively it produces blocky artifacts in the predicted depth map, especially for outdoor scenes with distant structures such as sky or mountains, where depth is unbounded and thus not well-defined.
As such, we retain the early low-resolution convolutional layers in DPT since they are computationally inexpensive.

\paragraph{Camera.}

The cameras $(\bg_1,\dots,\bg_N)$ are predicted jointly by applying a lightweight transformer to the camera tokens and registers $\{(\zcamera_i,\zscene_i)\}_{i=1}^N$, followed by an MLP on each updated camera token.
Unlike VGGT, our camera head predicts camera parameters in a single pass, without iterative refinement.

\subsection{Training Losses}%
\label{sec:training-losses}

In VGGT, we found it beneficial to predict redundant dense heads (\eg, point maps and tracks), but doing so is expensive during training.
Instead, \method contains a single dense head for depth prediction.
Although the model does not \emph{directly} predict point maps and tracks, we \emph{still} supervise these quantities through corresponding losses.
We found this yields nearly the same performance as using multiple dense prediction heads while saving a significant amount of memory.
We thus optimize the loss:
{\small
\begin{equation}
\label{eq:training-loss}
\mathcal{L} =
\lambda_{\text{cam}}   \mathcal{L}_{\text{cam}} +
\lambda_{\text{depth}} \mathcal{L}_{\text{depth}} +
\lambda_{\text{point}} \mathcal{L}_{\text{point}} +
\lambda_{\text{match}} \mathcal{L}_{\text{match}}
\end{equation}}%
where $\lambda_{\text{cam}}$, $\lambda_{\text{depth}}$, $\lambda_{\text{point}}$, and $\lambda_{\text{match}}$ are weights.

\paragraph{Camera loss.}

The camera loss
$
\mathcal{L}_\text{cam} =
\sum_{i=1}^N \left| \hat{\bg}_i - \bg_i \right|
$
compares the predicted cameras $\hat{\bg}_i$ to the ground-truth ones $\bg_i$ using an $\ell_1$ objective, which we found to be more stable than the Huber loss used in VGGT\@.

\paragraph{Depth loss.}

Following VGGT, the depth loss uses aleatoric uncertainty and a gradient consistency term.
Additionally, we account for the relative scale with respect to the ground truth.
Therefore, we have:
$
\mathcal{L}_{\text{depth}} =
\sum_{i=1}^{N}
\left[
\left \|
c_i^{D} \odot \left( 1 + D_i^{-1} \right) \odot e_i \right \|
+
\left \| c_i^{D} \odot \nabla e_i \right \|
\right] - \alpha \sum_{i=1}^{N} \log c_i^{D}
$, where $e_i = \hat{D}_i - D_i$,  $c_i^{D}$ is the predicted uncertainty map, and $\odot$ denotes element-wise product.

\paragraph{Point loss.}

Point maps assign to each pixel the coordinates of the corresponding 3D point in the frame of the reference camera.
The point maps can thus be inferred from the depth maps and the camera parameters via unprojection.
Accordingly, our point loss $\mathcal{L}_{\text{point}}$ is the same as the depth loss $\mathcal{L}_{\text{depth}}$ up to replacing the residuals with
$
e_i = \pi^{-1}(\hat{D}_i, \hat{\bg}_i) - P_i
$,
where $\pi^{-1}$ denotes unprojection and $P_i$ is a point map.

\paragraph{Matching loss.}

The matching loss $\mathcal{L}_{\text{match}}$ is applied to the tokens output by the last attention layer.
It pulls together features of positive token pairs (corresponding to the same 3D location) and pushes apart negative pairs:
$
\mathcal{L}_{\text{match}} = \mathbb{E}_{\text{pos}}\!\big[-\log \sigma(s)\big] + \mathbb{E}_{\text{neg}}\!\big[-\log(1-\sigma(s))\big]
$,
where $s$ is the cosine similarity between $\ell_2$-normalized tokens, $\sigma$ is the sigmoid function, and $\mathbb{E}$ denotes averaging over positive and negative pairs, \ie, a weighted binary cross-entropy.
Details of how to construct the pairs are provided in the supplement.

\subsection{Dynamic Reconstruction}%
\label{sec:dynamic-reconstruction}

\method supports reconstructing dynamic scenes, which, among other benefits, unlocks orders-of-magnitude more training data (since most videos contain motion).
Dynamic reconstruction requires a statistical prior that constrains movement.
Classical non-rigid structure from motion~\cite{bregler00recovering, dai14a-simple,taylor10non-rigid} imposes hand-designed low-rank or local-rigidity constraints, which can be brittle and limited in generality.
A data-driven model like \method has the potential to learn a better prior from data.
However, the output representation determines how camera motion and scene motion are coupled. 
In particular, some recent methods~\cite{wang24dust3r:} use point maps to represent the scene and recover camera parameters from them.
This works well for static scenes, but in dynamic scenes it requires either segmenting out moving pixels, as in MonST3R~\cite{zhang24monst3r:}, or introducing extensions such as dynamic point maps~\cite{sucar25dynamic,sucar26v-dpm}.
Another option is to predict depth and ray maps~\cite{lin25depth}.
However, ray maps add an expensive dense output and can entangle camera information with pixel-wise appearance changes.
For example, a stationary camera observing a dancer has large motion but fixed camera parameters.
Therefore, as stated above, we predict only depth maps and camera parameters, and avoid explicit dynamic outputs such as motion masks.

\subsection{Self-supervised Training}%
\label{sec:self-supervised-training}

Inspired by common practice in 2D vision~\cite{he20momentum, caron21emerging,oquab24dinov2:, simeoni25dinov3}, we use a teacher-student strategy for self-supervised learning with unlabeled videos.
Specifically, we maintain a \emph{student} network that is updated by gradient descent and a \emph{teacher} network that is updated only via an exponential moving average of the student network.
Both networks are initialized from the \method checkpoint trained on supervised data.
Given a video sequence, we feed the same set of frames to both networks but apply independent stochastic augmentations, including color jittering and blurring, random $90^\circ$ rotations, random patch masking, and random frame reordering (which affects the selection of the reference frame).
After restoring both streams to a common order, we require the student to match the teacher in two ways.
An $\ell_2$ feature-matching loss aligns the student's tokens with the teacher's across multiple layers.
Regression losses supervise the camera and depth.
To prevent collapse, the camera and depth heads are \emph{frozen} during self-supervision.
The teacher's parameters are updated as $\theta^{\mathrm{T}} \leftarrow m\,\theta^{\mathrm{T}} + (1-m)\,\theta^{\mathrm{S}}$ instead of gradient descent.
This distillation scheme thus enforces invariance to appearance changes and frame order, enabling effective learning from millions of unlabeled videos.

\subsection{Training Data}%
\label{sec:data}

\begin{table*}[htbp]
\caption{\textbf{Camera Pose Estimation} across static and dynamic benchmarks.
The metric AUC is higher-is-better. 
Feed-forward models (DA3, PI3, VGGT) are robust across datasets, but still lag behind the dynamic optimization-based method MegaSaM at strict thresholds on Sintel (\eg, AUC@$3^\circ$ of $16.2$ \vs $22.5$). 
In contrast, dynamic optimization methods (MegaSaM, MonST3R) degrade on wide-baseline scenes (\eg, AUC@$30^\circ$ of $38.1$ \vs $86.4$ on ETH3D). 
Our method, however, substantially advances the state of the art across all scenarios, \eg, improving AUC@$3^\circ$ from $22.5$ to $40.0$ on Sintel, with a $77\%$ relative improvement. Results updated, see GitHub for details.}%
\label{tab:camera}
\centering
\scriptsize
\setlength{\tabcolsep}{6pt}
\resizebox{\textwidth}{!}{%
\begin{tabular}{c *{6}{cc}}
\toprule
\multirow{3}{*}{\textbf{Method}}
& \multicolumn{6}{c}{\textbf{Static Scenes}}
& \multicolumn{6}{c}{\textbf{Dynamic Scenes}} \\
\cmidrule(lr){2-7}\cmidrule(lr){8-13}
& \multicolumn{2}{c}{\textbf{7 Scenes}}
& \multicolumn{2}{c}{\textbf{NRGBD}}
& \multicolumn{2}{c}{\textbf{ETH3D}}
& \multicolumn{2}{c}{\textbf{DyCheck}}
& \multicolumn{2}{c}{\textbf{Sintel}}
& \multicolumn{2}{c}{\textbf{TUM\text{-}Dynamic}} \\
& AUC@3° & AUC@30°
& AUC@3° & AUC@30°
& AUC@3° & AUC@30°
& AUC@3° & AUC@30°
& AUC@3° & AUC@30°
& AUC@3° & AUC@30° \\
\midrule
{MonST3R} 
& 9.0 & 68.3 & 13.9 & 79.7 & 1.7 & 14.3 
& 11.5 & 45.4 & 4.3 & 45.8 & 7.7 & 48.5 \\
MapAnything & 5.8 & 61.4 & 35.2 & 88.9 & 13.2 & 51.0 & 6.1 & 60.3 & 2.9 & 31.6 & 4.3 & 40.2 \\
{MegaSaM} 
& 10.6 & 71.8 & 17.2 & 83.1 & 5.9 & 38.1 
& 26.8 & 53.1 & 22.5 & 58.3 & 15.4 & 59.0 \\
{VGGT}    
& 10.9 & 74.4 & 81.7 & 97.7 & 18.8 & 62.1 
& 21.0 & 78.7 & 15.0 & 50.0 & 16.6 & 61.2 \\
{PI3}     
& 13.3 & 77.0 & 83.8 & 98.2 & 35.3 & 79.6
& 23.3 & 81.0 & 14.8 & 53.5 & 16.1 & 59.2 \\
{DA3} & 18.7 & 78.2 & 86.4 & 98.4 & 46.1 & 87.0 & 32.1 & 83.9 & 16.2 & 52.7 & 20.8 & 62.7 \\
\midrule
{Ours-1B} 
& \underline{27.4} & \underline{82.4}
& \underline{90.1} & \underline{98.9}
& \underline{51.5} & \underline{88.9}
& \underline{35.9} & \underline{86.5}
& \underline{35.5} & \underline{73.4}
& \underline{30.0} & \underline{82.8} \\
{Ours-10B} 
& \textbf{36.4} & \textbf{88.2} 
& \textbf{92.5} & \textbf{99.1} 
& \textbf{56.3} & \textbf{90.4} 
& \textbf{43.7} & \textbf{90.9} 
& \textbf{40.0} & \textbf{79.1} 
& \textbf{36.4} & \textbf{87.5} \\
\bottomrule
\end{tabular}%
}
\vspace{-1em}
\end{table*}

\begin{table*}[htbp]
\caption{\textbf{Depth Estimation} across static and dynamic benchmarks.
The metric \(\delta_{1.25}\) denotes the percentage of predicted depths within a factor of the ground truth (higher is better), and AbsRel is the mean absolute relative error (lower is better). }%
\label{tab:depth}
\centering
\tiny
\setlength{\tabcolsep}{6pt}
\resizebox{\textwidth}{!}{%
\begin{tabular}{c *{6}{cc}}
\toprule
\multirow{3}{*}{\textbf{Method}}
& \multicolumn{6}{c}{\textbf{Static Scenes}}
& \multicolumn{6}{c}{\textbf{Dynamic Scenes}} \\
\cmidrule(lr){2-7}\cmidrule(lr){8-13}
& \multicolumn{2}{c}{\textbf{7 Scenes}}
& \multicolumn{2}{c}{\textbf{NRGBD}}
& \multicolumn{2}{c}{\textbf{ETH3D}}
& \multicolumn{2}{c}{\textbf{DyCheck}}
& \multicolumn{2}{c}{\textbf{Sintel}}
& \multicolumn{2}{c}{\textbf{TUM\text{-}Dynamic}} \\
& \(\delta_{1.25}\) & {AbsRel}
& \(\delta_{1.25}\) & {AbsRel}
& \(\delta_{1.25}\) & {AbsRel}
& \(\delta_{1.25}\) & {AbsRel}
& \(\delta_{1.25}\) & {AbsRel}
& \(\delta_{1.25}\) & {AbsRel} \\
\midrule
{MonST3R} & 92.4 & 0.075 & 98.4 & 0.030 & 95.8 & 0.056 & 93.3 & 0.068 & 71.9 & 0.263 & 85.0 & 0.148 \\
MapAnything & 92.9 & 0.070 & 98.7 & 0.022 & 96.3 & 0.035 & 97.0 & 0.049 & 72.5 & 0.251 & 93.1 & 0.052 \\
{MegaSaM} & \underline{93.8} & 0.065 & 96.2 & 0.057 & 94.8 & 0.083 & 97.4 & 0.042 & 74.1 & 0.207 & 92.9 & 0.083 \\
{VGGT}    & 91.9 & 0.073 & 99.1 & 0.019 & 97.4 & 0.036 & 95.2 & 0.055 & 79.2 & 0.189 & 92.2 & 0.064 \\
{PI3}     & 92.8 & 0.068 & 99.2 & 0.011 & {99.6} & 0.016 & 97.4 & 0.041 & 82.5 & 0.144 & 95.5 & 0.046 \\
{DA3} & 93.0 & 0.063 & 99.5 & \underline{0.010} & {99.6} & 0.015 & 97.7 & \underline{0.039} & 86.1 & 0.118 & 94.3 & 0.049 \\
\midrule
{Ours-1B} 
& 93.7 & \underline{0.060}
& \underline{99.6} & 0.011
& \underline{99.8} & \underline{0.012}
& \underline{97.9} & \underline{0.039}
& \underline{89.9} & \underline{0.094}
& \underline{97.7} & \underline{0.040} \\
{Ours-10B} 
& \textbf{96.3} & \textbf{0.050} 
& \textbf{99.7} & \textbf{0.007} 
& \textbf{99.8} & \textbf{0.009} 
& \textbf{98.7} & \textbf{0.030} 
& \textbf{93.5} & \textbf{0.081} 
& \textbf{98.3} & \textbf{0.035} \\
\bottomrule
\end{tabular}%
}%
\vspace{-10pt}
\end{table*}

An important aspect of \method is to scale the training data, which we achieve by combining a large number of publicly available datasets (\cref{sec:data-sources}) with a new annotation pipeline that we develop to handle dynamic content in off-the-shelf videos (\cref{sec:annotation-pipeline}).

\subsubsection{Data Sources}%
\label{sec:data-sources}

We first collect several publicly available datasets:
Aria series~\cite{pan23aria},
Bedlam~\cite{black23bedlam:},
BEHAVIOR-1K~\cite{li24behavior-1k:},
Co3Dv2~\cite{reizenstein21common},
uCo3D~\cite{liu25uco3d},
DL3DV~\cite{ling23dl3dv-10k:},
Dynamic Replica~\cite{karaev23dynamicstereo},
EDEN~\cite{le21eden:},
EFM3D~\cite{straub24efm3d:},
HOT3D~\cite{banerjee24introducing},
Habitat~\cite{habitat19iccv},
Hypersim~\cite{roberts21hypersim:},
Mapfree~\cite{arnold22map-free},
Mapillary Metropolis~\cite{mapillary25metropolis},
MPSD~\cite{antequera20mapillary},
Megadepth~\cite{li18megadepth:},
Megasynth~\cite{jiang25megasynth:},
Mid-Air~\cite{fonder19mid-air:},
Mvssynth~\cite{huang18deepmvs:},
ParallelDomain-4D~\cite{hoorick24generative},
Replica~\cite{straub19the-replica},
SAIL-VOS~\cite{hu21sail-vos},
ScanNet Series~\cite{dai17scannet:, yeshwanth23scannet:},
TartanAirV2~\cite{wang20tartanair:},
TartanGround~\cite{patel25tartanground:},
Taskonomy~\cite{zamir18taskonomy:},
UnrealStereo4K~\cite{tosi21smd-nets:},
Virtual KITTI~\cite{cabon20virtual},
Waymo~\cite{sun20scalability}, and
WildRGBD~\cite{xia24rgbd}.
We exclude Kubric~\cite{greff22kubric:} and PointOdyssey~\cite{zheng23pointodyssey:} used by VGGT because their background geometry is fake and yields invalid depth.
Additionally, we use several internal datasets, which include artist-created object assets, rigid and dynamic synthetic environments, real-world device captures, and related sources.
For non-synthetic datasets (\eg, those annotated by SfM pipelines), we remove noisy depth values via a multi-view consistency check and discard sequences with too few valid depth pixels.
In total, these datasets contain approximately $3$M sequences, each containing between $10$ and $20{,}000$ images.

\subsubsection{Data Annotation Pipeline}%
\label{sec:annotation-pipeline}

To further expand our training data, we built a large internal video collection of roughly $40$M Internet-style videos.
We first assess each video for suitability for reconstruction, \eg filtering out clips with large watermarks or abrupt shot changes.
Videos that pass this check are used for self-supervised training as described in \cref{sec:self-supervised-training}.
We then introduce a new pipeline to annotate videos with camera parameters and depth maps, for both static and dynamic scenes.
We prioritize annotation quality over quantity and aggressively reject low-quality data.
We also discard depth annotations in regions that are likely to be dynamic.
This may exclude extreme camera motions or highly dynamic scenes, but these are still well represented in the synthetic datasets.
Overall, we obtain a collection of about $200$K dynamic scenes and $600$K static scenes with high-quality camera and depth annotations.

\paragraph{VLM pre-filtering.}

We prompt a Vision-Language Model (VLM) to discard videos that are unlikely to be reconstructed using multi-view geometry.
The VLM classifies $50\%$ of the clips as too difficult to reconstruct, \eg, because they contain multiple clips, extreme motion blur, or heavy overlays or watermarks.
It also classifies $40\%$ of them as reconstructible, but potentially with low accuracy due to insufficient parallax, lack of non-repetitive texture, etc.
The remaining $10\%$ of videos go to the next stage.
In the same pass, the VLM extracts metadata, such as whether the scene appears dynamic, for use by later stages.

\paragraph{Dynamic mask extraction.}

We use Grounding DINO~\cite{liu24grounding} to detect bounding boxes of potentially movable object categories, such as people and cars.
These regions are then excluded from matching, tracking, and verification.

\paragraph{Feature matching and tracking.}

We extract matches and tracks across frames with an ensemble of methods, using
SIFT~\cite{lowe99object},
SuperPoint \& SuperGlue~\cite{daniel18superpoint:, sarlin20superglue:},
ALIKED \& LightGlue~\cite{zhao23aliked:, lindenberger23lightglue:}, and
VGGSfM Tracker~\cite{wang24vggsfm:}.
Matches within dynamic regions are discarded.

\paragraph{Reconstruction and filtering.}

We use the original VGGT to initialize camera parameters when RANSAC-based essential matrix estimation yields too few inliers, and then run COLMAP~\cite{schonberger16structure-from-motion} for iterative bundle adjustment and filtering based on the correspondences computed above.
For successful reconstructions, we discard sequences that fail heuristic checks, \eg, an image registration ratio $<99.5\%$, a field of view outside $[30^\circ,120^\circ]$, or a distortion ratio $>0.1$.
These criteria aggressively remove cases with degenerate motion or extreme zoom.
Then, we estimate per-frame dense depth maps using patch-based multi-view stereo~\cite{schonberger16pixelwise}.

\paragraph{Multi-view consistency.}

For each frame, we unproject the depth map to 3D, reproject the points into other views, and compare them with the depths there.
Pixels that satisfy this cross-view consistency check are marked valid.
We discard sequences with fewer than $5\%$ of pixels with valid depth, which typically, though not always, indicates low-quality cameras.

\paragraph{Supervised geometric filtering.}

Finally, we obtain cameras, depths, and valid masks for every sequence.
We use handcrafted features, such as camera-up-vector consistency, parallax angle, and trajectory smoothness, to describe each sequence.
We hand-annotated 500 static and 500 dynamic sequences, respectively, to train a classifier to remove low-quality reconstructions.
The classifier is an ensemble of XGBoost~\cite{chen16xgboost:}, random forests~\cite{breiman01random}, and CatBoost~\cite{prokhorenkova18catboost:}.

\section{Experiments}%
\label{sec:experiments}

Here, we provide additional details about our framework, benchmark it against state-of-the-art rigid and dynamic reconstruction methods, and ablate key design choices.

\subsection{Implementation Details}

We focused on four model variants: $200$M, $500$M, $1$B, and $10$B parameters, with $12/12/24/16$ alternating-attention blocks and hidden sizes $384/768/1024/4096$, respectively.
The vision transformer is initialized from DINOv3~\cite{simeoni25dinov3} and is not frozen during training.
Each block contains one global attention (or register attention) layer and one frame-wise attention layer.
Optimization uses AdamW for $240$K iterations, with $160$K supervised, $50$K self-supervised, and a final $30$K supervised stage.
The learning rate follows a linear warm-up over $5\%$ of training and cosine decay for the remaining $95\%$, with a peak value of $2\times10^{-4}$ for supervised training and $1\times10^{-4}$ for self-supervised training.
For each batch, the number of frames is drawn uniformly from $[1,24]$.
We augment images by randomly varying the aspect ratio within \([0.33,1.33]\), keeping the image area approximately $512 \times 512$ pixels, and applying color jittering, grayscale conversion, and random patch masking.
Training used $128$ $96$GB H100 GPUs, bfloat16 mixed precision, gradient checkpointing, and FSDP.

\subsection{Benchmarking}%
\label{sec:benchmarking}

\begin{figure*}
\centering
\includegraphics[width=1.0\textwidth]{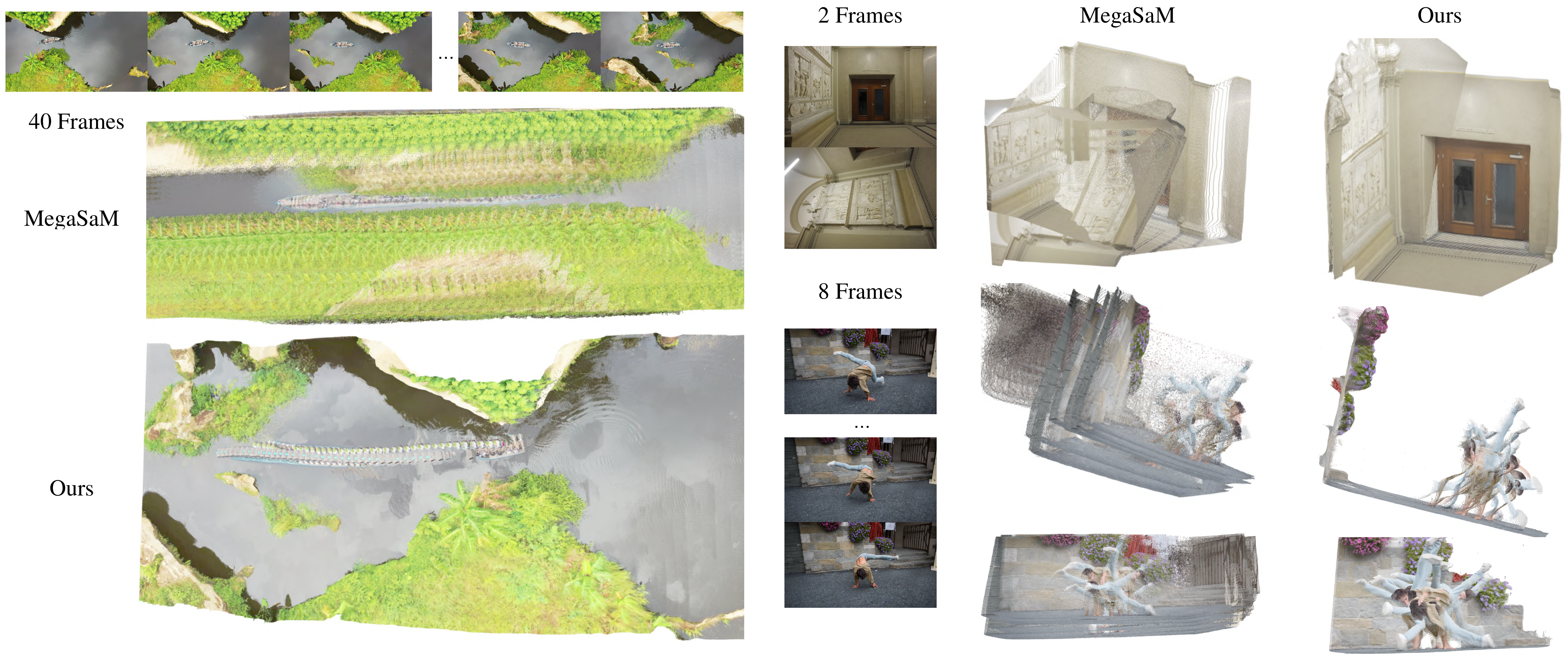}
\caption{\textbf{Qualitative Comparison to MegaSaM.} 
MegaSaM often suffers from geometric inconsistencies. 
In contrast, our method produces globally consistent reconstructions across sparse, dynamic, and aerial scenarios. 
This is especially evident in the aerial scene in the left column, where MegaSaM exhibits severe geometric drift and texture smearing, resulting in repeated patterns and a distorted global layout.
Note that the second frame in the top right example is not upright, which makes this case particularly challenging. 
}%
\label{fig:qual_supp}
\end{figure*}

\paragraph{Quantitative Comparison.}

We compare \method with recent approaches:
(i) feed-forward reconstruction models and (ii) optimization-based dynamic reconstruction methods.
We evaluate on three static datasets (7 Scenes~\cite{shotton13scene}, NRGBD~\cite{azinovic22neural}, and ETH3D~\cite{schops17a-multi-view}) and three dynamic datasets (DyCheck~\cite{gao22monocular}, Sintel~\cite{butler12a-naturalistic}, and TUM-Dynamic~\cite{sturm12a-benchmark}).
For each scene or sequence, we randomly sample $10$ frames.
We use the original released models for all methods.
For DA3, we use its largest variant, Giant (1B parameters).

Following~\cite{wang25vggt}, we report the standard AUC for camera pose estimation (higher is better), computed as the area under the curve of the fraction of image pairs whose relative rotation and translation errors fall below an angular threshold (\eg, $3^\circ$, $30^\circ$).
As shown in \cref{tab:camera}, feed-forward models generally exhibit strong performance on static benchmarks and at more relaxed thresholds, while optimization-based, dynamic-aware MegaSaM is more competitive on challenging dynamic sequences such as Sintel but degrades on wide-baseline or low-texture scenes.
In contrast, our models consistently outperform all baselines across both static and dynamic datasets and at both strict and relaxed thresholds.

We also evaluate the accuracy of the predicted depths using absolute relative error (AbsRel; lower is better) and $\delta_{1.25}$ (higher is better), which measures the percentage of pixels for which the ratio of the predicted depth to the ground-truth depth is within a specified threshold.
As shown in \cref{tab:depth}, our models outperform the baselines in the static benchmarks, further lowering AbsRel on datasets where existing methods perform strongly, such as ETH3D, and even more so in dynamic scenes, where they reduce depth errors and increase $\delta_{1.25}$ (\eg, on Sintel, improving $\delta_{1.25}$ from $86.1$ to $93.5$ and AbsRel from $0.118$ to $0.081$).

The larger 10B variant consistently outperforms the 1B model, indicating that scaling up the reconstruction model directly benefits both camera and depth accuracy.

\paragraph{Qualitative Results.}

\Cref{fig:qual} illustrates our results on static and dynamic scenes, including traffic, human motion, natural landscapes, and underwater environments.

\begin{figure*}
\centering
\includegraphics[width=\textwidth]{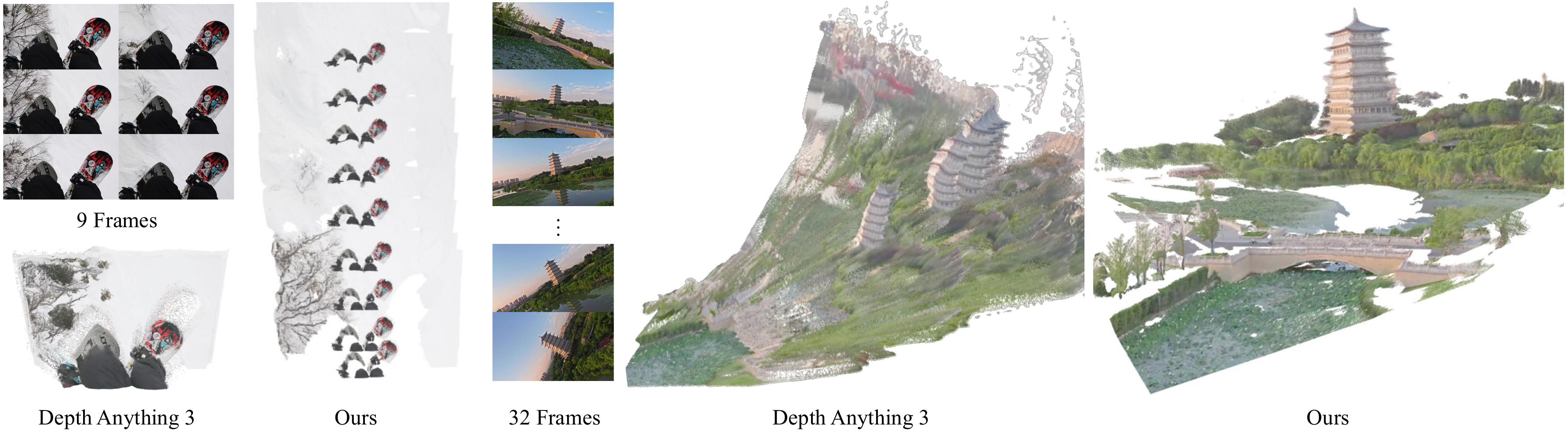}
\caption{\textbf{Qualitative Comparison to Depth Anything 3.}
\emph{Left:} a snow lift sequence over a snow-covered field, where the repetitive terrain misleads DA3 into estimating little to no camera motion.
Our method recovers the correct camera trajectory.
\emph{Right:} a drone sequence in which the camera rolls while flying toward a tower.
Under this strong viewpoint change, DA3 produces severe ghosting and duplicated tower structures, whereas our reconstruction remains globally consistent.}%
\label{fig:qual_da3}
\end{figure*}

\begin{figure}[htbp]
\centering
\includegraphics[width=\linewidth]{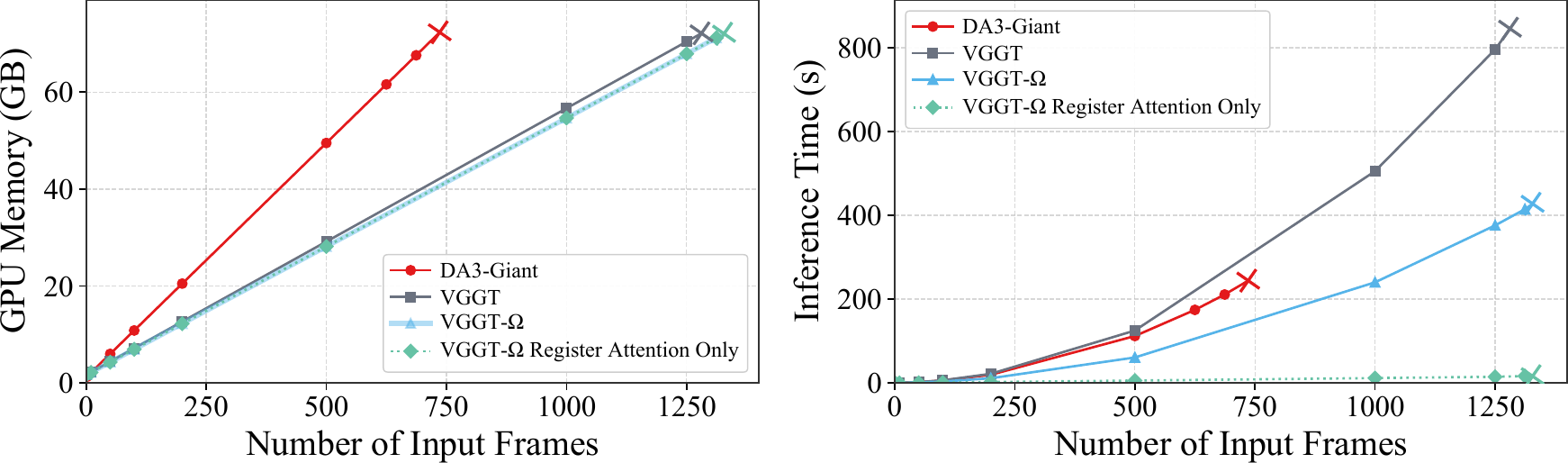}
\caption{\textbf{Memory and Speed Comparison.}
We compare the inference memory usage and runtime of VGGT, DA3, and \method on an 80GB A100 GPU with flash attention v2 enabled.
For VGGT, we first address an implementation detail which improves its efficiency (see text).
Cross markers indicate the first setting at which each method runs out of memory.
}%
\label{fig:mem_speed}
\end{figure}

We further compare \method with DA3 and MegaSaM on challenging cases in \cref{fig:qual_da3,fig:qual_supp}.
In \cref{fig:qual_da3}, DA3 struggles with repeated textures, estimating little to no camera motion in the snow lift sequence, and with strong camera roll, reconstructing the tower several times in the drone sequence.
MegaSaM can also break down in sparse indoor scenes with substantial camera roll (top right) or textureless walls (bottom right), producing disjoint structures or misaligned planes, while certain aerial sequences (left) also challenge its pose estimation.
In contrast, \method's reconstructions are globally consistent, likely due to the strength of the geometric priors learned from diverse data.

\paragraph{Inference Memory and Speed.}

We now compare the \emph{inference} memory and speed of \method with VGGT and DA3\@ (note that this is quite different from the training efficiency discussed in \cref{sec:architecture}).

For these comparisons, we first correct the original VGGT implementation.
That model caches intermediate tensors from all $24$ layers during inference.
Although these cached tensors are useful during training, the prediction heads only require intermediate features from four layers at test time.
We thus avoid caching unnecessary information, which substantially reduces inference memory usage.

With this correction in place, we compare VGGT, DA3, and \method in \cref{fig:mem_speed}. 
For a fair comparison, all measurements are conducted on a single 80GB A100 GPU using PyTorch \textit{scaled\_dot\_product\_attention}, with a flash attention v2 backend. 
We use similar input resolutions matched to each model's patch size: $518 \times 336$ for VGGT and DA3 with $14$-pixel patches, and $512 \times 336$ for \method with $16$-pixel patches.
We increase the number of input frames until each method runs out of memory, as indicated by the cross markers in \cref{fig:mem_speed}.

As for memory efficiency, VGGT (with the correction) and \method are similar, and can process more than $1,\!000$ frames on a single A100 GPU\@.
As for speed, \method is faster, primarily due to using DINOv3, with a patch size of $16$, instead of DINOv2, with a patch size $14$, which reduces the number of image tokens by about $25\%$.
In addition, \method replaces $25\%$ of the global attention layers with register attention by default, reducing FLOPs and yielding a $20$--$25\%$ speedup.
Overall, VGGT and \method scale more favorably than DA3 in both memory and runtime, \eg, DA3 runs out of memory at around $750$ frames in our testing, whereas VGGT and \method can handle around $1250$ frames.
An even more aggressive variant of \method that replaces all global attention layers with register attention can further improve speed, reducing the runtime on $1000$ frames from $240.2$ seconds to $11.7$ seconds, albeit at the cost of lower reconstruction accuracy (\cref{sec:discussion}).

While perhaps surprising, we stress that, during inference, the main benefit of removing global attention is speed, not memory.
As shown in \cref{fig:mem_speed}, replacing all global attention layers with register attention produces a large speedup, while peak inference memory remains almost unchanged. 
The speedup is because register attention avoids computing interactions between all pairs of image tokens, unlike global attention.
However, with PyTorch \textit{scaled\_dot\_product\_attention} using flash attention v2, the full attention matrix is not explicitly materialized in memory. 
Instead, the kernel computes attention in a tiled streaming manner, updates the softmax normalization online, and directly accumulates the output.
Peak memory is instead dominated by tensors whose size is proportional to the number of frames and image tokens, such as frame-attention activations or feed-forward intermediates.
This explains why memory grows approximately linearly with the number of frames in \cref{fig:mem_speed}.

\subsection{Ablation Studies}

Unless otherwise specified, ablations use the 1B model variant trained on $2$M sequences with $64$ GPUs for $150$k supervised steps.
To jointly assess camera and depth accuracy, we unproject depth maps into 3D using the cameras and compute the $\ell_2$ distance between the predicted and ground-truth points, which we call \emph{point error}.
We use point error rather than Chamfer distance because nearest-neighbor matching over unordered point sets can be dominated by large surface regions such as walls and floors.
All the models are trained on approximately the same number of tokens.

\paragraph{Model and data size.}

We observe that scaling either the model or the data consistently improves performance, as shown in \cref{fig:scale}.
Increasing the number of training sequences in $10\times$ steps yields a monotonic drop in point error, from $0.275$ to $0.073$.
Overall, the shape of both curves suggests that power laws might characterize scaling in this class of models.

\paragraph{Register attention.}

A variant that uses only global attention layers achieves a point error of $0.071$.
Replacing 25\% of the global attention layers with register attention yields performance nearly identical to the original ($0.073$).

\paragraph{Multi-task learning.}

Removing the point and matching losses increases the point error from $0.073$ to $0.078$.
For reference, adopting VGGT's original multi-head, multi-task setup achieves $0.070$ but requires multiple dense heads, making scaling difficult.

\paragraph{Self-supervised training.}

Replacing $10\%$ of training steps from supervised training with self-supervised training slightly reduces point error from $0.073$ to $0.070$.
This performance gain stems from training on unlabeled data, which is more diverse.
We also observe improved out-of-distribution generalization.

\paragraph{Annotation quality.}

To validate the quality of the pseudo ground truth produced by our annotation pipeline, we compare it against MegaSaM~\cite{li25megasam:} on Sintel, which provides synthetic camera and depth ground truth.
For a fair comparison, we evaluate only the sequences and pixels that satisfy both our filtering criteria and MegaSaM's validation, excluding $8/23$ sequences and all dynamic regions.
Under this protocol, our pipeline achieves $96.4\%$ AUC@$30^\circ$ for camera pose and $99.3\%$ $\delta_{1.25}$ for depth, compared with $62.1\%$ and $77.2\%$ for MegaSaM, respectively, confirming the high quality of the resulting annotations.
Our goal in pseudo-label generation is not to maximize yield, but to retain only sequences and pixels that are very likely to be correct, as we find that a smaller set of highly accurate pseudo ground truth is more beneficial in practice than a larger but noisier collection.
Hence, the annotation pipeline is intentionally conservative: if a sequence is even mildly ambiguous, or if a pixel cannot be validated reliably, we prefer to discard it rather than risk injecting noisy supervision.

\subsection{Applications of Registers}%
\label{sec:applications}

\paragraph{Robotics.} Recent work has explored the use of reconstruction models to improve spatial understanding in VLA systems~\cite{li25spatial, abouzeid25geoaware-vla:}.
We evaluate whether \method can be a plug-and-play geometric encoder for VLA\@.
Given the input images, we extract registers (scene tokens) from \method and concatenate them with the original OpenVLA-OFT input tokens~\cite{kim25fine-tuning}.
We then train OpenVLA-OFT using its standard protocol, while keeping \method fixed throughout.
As shown in \cref{tab:vla}, the geometry-aware registers consistently improve performance across all LIBERO tasks~\cite{liu23libero:}.

\begin{table}[htbp]
\centering
\caption{\textbf{Performance on the LIBERO benchmark.}
We freeze our pretrained model, feed the scene tokens as additional input to OpenVLA-OFT, and report the success rate (SR) (higher is better).
The clear gains validate the effectiveness of our scene tokens in aggregating spatial information.
}%
\label{tab:vla}
\scriptsize
\renewcommand{\tabcolsep}{4pt}
\begin{tabular}{cccccc}
\toprule
\multicolumn{1}{c}{Method}
& \begin{tabular}[c]{@{}c@{}}Spatial\\ SR (\%)\end{tabular}
& \begin{tabular}[c]{@{}c@{}}Object\\ SR (\%)\end{tabular}
& \begin{tabular}[c]{@{}c@{}}Goal\\ SR (\%)\end{tabular}
& \begin{tabular}[c]{@{}c@{}}Long\\ SR (\%)\end{tabular}
& \begin{tabular}[c]{@{}c@{}}Average\\ SR (\%)\end{tabular}
\\ \midrule
Diffusion Policy & 78.3 & 92.5 & 68.3 & 50.5 & 72.4 \\
TraceVLA         & 84.6 & 85.2 & 75.1 & 54.1 & 74.8 \\
Octo             & 78.9 & 85.7 & 84.6 & 51.1 & 75.1 \\
OpenVLA          & 84.7 & 88.4 & 79.2 & 53.7 & 76.5 \\
Dita             & 84.2 & 96.3 & 85.4 & 63.8 & 82.4 \\
CoT-VLA          & 87.5 & 91.6 & 87.6 & 69.0 & 83.9 \\
$\pi_0$-FAST     & 96.4 & 96.8 & 88.6 & 60.2 & 85.5 \\
$\pi_0$          & 96.8 & 98.8 & 95.8 & 85.2 & 94.2 \\
UniVLA           & 96.5 & 96.8 & 95.6 & 92.0 & 95.2 \\
OpenVLA-OFT      & 97.6 & 98.4 & 97.9 & 94.5 & 97.1 \\
\midrule
\begin{tabular}[c]{@{}c@{}}OpenVLA-OFT\\ + Our Frozen Scene Tokens\end{tabular} & \textbf{99.3} & \textbf{99.2} & \textbf{99.0} & \textbf{96.7} & \textbf{98.5} \\
\bottomrule
\end{tabular}
\end{table}

\paragraph{Language Alignment.}

\begin{figure}
\centering
\includegraphics[width=0.95\linewidth]{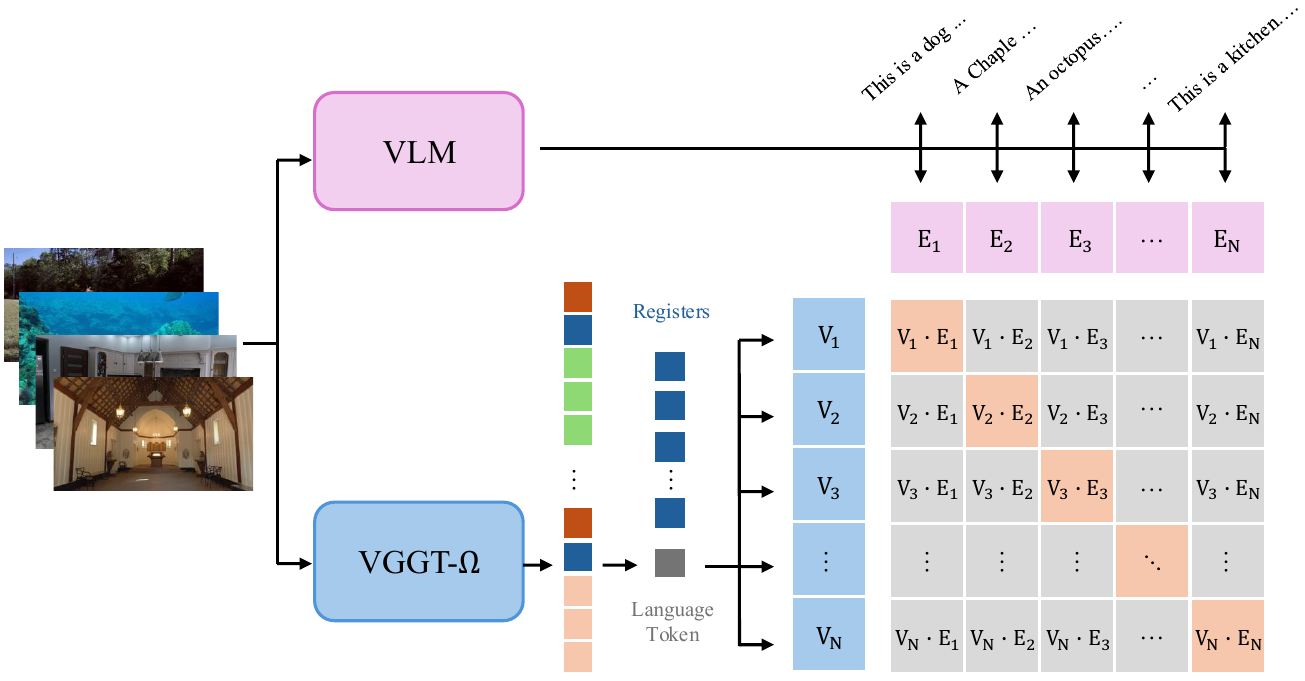}
\caption{\textbf{Language Alignment.} 
We conduct language alignment to verify whether the registers contain high-level information. 
For each sequence, a VLM describes the scene content, coarse layout, and appearance, and the generated text tokens are mean-pooled to form the language embedding. 
On the \method side, a small self-attention stack takes the registers and a learnable language token as input, and the output language token is projected to form the register-derived embedding. 
The two embeddings are optimized with a symmetric InfoNCE loss over all sequence-description pairs in the global batch.
}%
\label{fig:text_alignment}
\end{figure}

To further verify whether the registers contain high-level information, we investigate whether they can be aligned to natural language.
The procedure, illustrated in \cref{fig:text_alignment}, follows the spirit of CLIP-style contrastive alignment~\cite{radford21learning, zhai23sigmoid}.
However, to keep it as simple as possible, we use a well-trained \method, fix the language encoder and only fine-tune our model.

In detail, we use a VLM and \method to separately extract a global descriptor for a given image sequence and then align the two.
To extract this global descriptor, the VLM observes all input views and is prompted to describe the scene content, coarse layout, and appearance as a single coherent scene.
Then, the hidden states of the text tokens generated by the VLM are mean-pooled and $\ell_2$-normalized.
To obtain a global descriptor from \method, a small self-attention stack takes the registers and a new learnable language token (randomly initialized) as input.
The output language token is projected and $\ell_2$-normalized to produce the register-derived embedding from \method.

It is worth noting that the language token we introduced never directly observes image patch tokens, as it can only read out the registers.
Thus, a successful alignment indicates that the registers themselves carry scene-level information that can be matched to language.

We maximize the cosine similarity of the matched register-derived embeddings and VLM language embeddings while minimizing the similarity of mismatched pairs with a symmetric InfoNCE loss.
We fine-tune the models on the same image sequences as in the main reconstruction training.
As in distributed CLIP training, embeddings are gathered across GPUs so that the global batch provides in-batch negatives.
The VLM is frozen, and \method is fine-tuned end-to-end with a constant learning rate of $1\times10^{-5}$, rather than kept fixed as in the robotics experiment.

After only 10K iterations with a small learning rate, the model already transfers well to language retrieval.
For evaluation, we construct a benchmark of 100 manually curated internet videos spanning a diverse set of scenarios, including cooking, praying, car racing, and more.

Given the register-derived embedding of the aligned \method, we retrieve the paired language embedding for each candidate video using cosine similarity and report top-$K$ accuracy.
Using the VLM embedding employed during alignment, the model achieves $76.8\%$ top-1 accuracy and $97.0\%$ top-3 accuracy.
That is, the correct language description ranks among the top three candidates for almost all videos.
To test whether this alignment transfers beyond the exact training target, we replace the VLM embedding with a text-only LLM embedding, without any additional training.
Specifically, we prompt the VLM to generate a video description and feed only this description to the Qwen3 LLM~\cite{yang25qwen3}.
The model still obtains $47.5\%$ top-1 accuracy and $77.8\%$ top-3 accuracy in this zero-shot transfer setting.

Overall, these results show that the registers carry high-level information, likely quite semantic, that can be aligned with language space.
At the same time, there is no degradation on the geometric tasks after alignment fine-tuning.
The example text description and the prompt are provided in the supplementary materials.
This may indicate that representations learned by a strong geometry model can align naturally with those of other modalities, such as language.
This observation is consistent with the platonic representation hypothesis~\cite{huh24position:,han25learning}, which suggests that sufficiently capable models trained on different modalities tend to converge toward a shared representation space.

\paragraph{General Applications.}

Registers provide a general mechanism for extracting both sequence-level and frame-level representations from the model.
This design naturally extends to additional prediction tasks by introducing task-specific learnable tokens.
For sequence-level predictions, a learnable token can be concatenated to the register tokens and decoded into the desired output, similar to the language token discussed above.
For frame-level predictions, the same token can be replicated across frames, concatenated with the corresponding registers, and used to produce per-frame estimates.
We have conducted preliminary experiments on several such tasks, including metric scale estimation, gravity direction estimation, and human presence detection, and observed promising results.
We leave the exploration of these possibilities to the community.
From this perspective, a camera token can also be viewed as a register-like learnable token, with the main distinction that it is trained under direct camera supervision.

\section{Further Insights}%
\label{sec:discussion}

\begin{figure}
\centering
\includegraphics[width=0.95\linewidth]{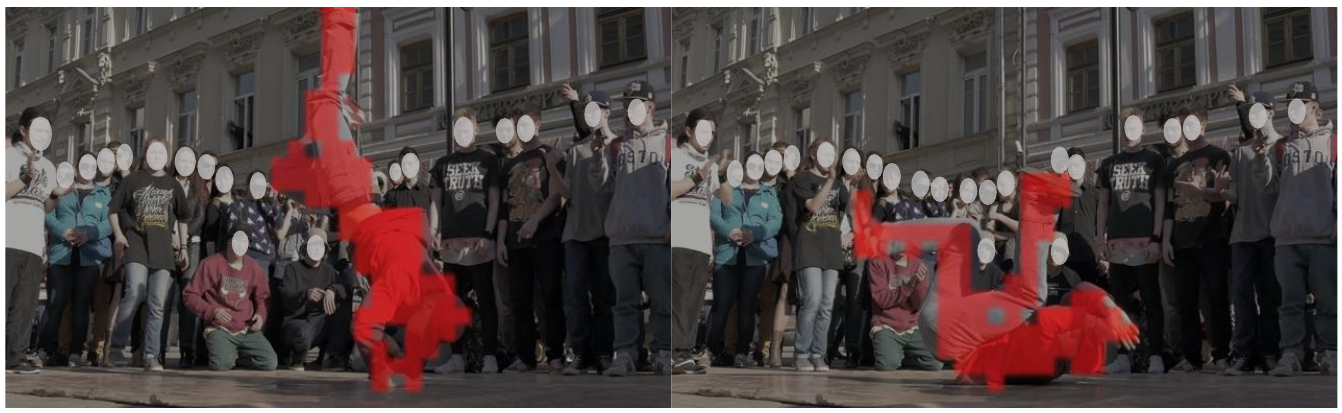}
\caption{\textbf{Motion-Aware Representations.}
We cluster PCA-reduced intermediate image tokens by $k$-means and observe that the resulting clusters consistently separate the moving dancer from the static crowd and background, with red indicating high-response regions.
This suggests that \method learns motion-aware representations from reconstruction training, without explicit motion supervision.
}%
\label{fig:motion}
\end{figure}

As we developed \method, we made several empirical observations that are not yet rigorously established but that we think are still useful to share with other researchers interested in developing similar models.

\paragraph{Model Souping: Where Is the Information Stored?}

To better understand the model's behavior, we use model souping~\cite{wortsman22model} as a probe.
Specifically, we fuse the public VGGT checkpoint into \method by directly averaging specific subsets of their weights, without any further training.
Perhaps surprisingly, although the two models use different architectures (\eg, VGGT uses DINOv2 with a patch size of 14, whereas \method uses DINOv3 with a patch size of 16 and register attention), the averaged models can still produce visually reasonable reconstructions.

By fusing different subsets of the weights, we can gain insights into where different types of information are stored in the model.
Our first observation is that depth and field-of-view information is largely stored in the FFNs of the attention blocks, primarily in the frame-wise attention blocks and, to a lesser extent, in the global attention blocks.
This is consistent with the observations in the language community~\cite{geva21transformer,dai22knowledge,meng22locating}.
For example, as discussed in~\cref{sec:data_quality}, the model can memorize idiosyncratic noisy signals in the training data, such as treating humans as part of the ground in certain scenarios.
When we fuse the FFN weights of VGGT's frame-wise attention blocks into \method with a 50\%-50\% average, these errors are resolved.
In contrast, fusing the Q/K/V projection weights does not produce the same effect.
Camera extrinsic information appears to be encoded at a higher level and is not controlled solely by the FFN weights.
Finally, the ability to generalize to a variable number of input frames is also closely related to the frame-wise attention blocks.

\paragraph{Motion Awareness.}

Since the model can reconstruct dynamic sequences with large non-camera motion, we ask whether it learns to localize the moving regions.
To this end, we analyze the feature distributions of its intermediate tokens.
Specifically, we normalize intermediate image tokens across space and time, reduce their dimensionality with PCA, and group them using $k$-means.
No labels, optical flow, or learned probes are used at any stage.

As shown in \cref{fig:motion}, one PCA cluster consistently tracks the moving dancer across all frames while the static crowd and background remain in the other clusters.
This suggests that the ability to discriminate motion emerges automatically as a byproduct of the reconstruction objective, without explicit motion supervision.
Note also that the model is never given the temporal order of the input frames, either during training or inference.

We further examine how this signal evolves across layers.
Early layers (\eg, layer 4 in \cref{fig:motion}) produce the cleanest motion segmentation, isolating the dancer with minimal background response.
Middle layers, such as layer 13, retain a weaker but still discernible motion signal.
At the deepest layers (\eg, layer 23), the same clustering procedure highlights all people in the scene, indicating that the representations become increasingly global and semantic.

\paragraph{Normalizing Predictions.}

In VGGT, we discussed whether predictions should be normalized online to a unit coordinate space.
We further analyze this question here.
Consistent with the observation in VGGT, once the model has converged, we observe no difference in quantitative performance between training with and without prediction normalization.
The main benefit of normalization is qualitative: the final point clouds appear more spatially well spread. 
The drawback is that the optimization becomes less stable, with a steeper learning curve and a greater need for careful hyperparameter tuning to avoid gradient explosion.
If prediction normalization is necessary, we suggest initializing from a pretrained model to stabilize training, as in PI3.

\paragraph{DINO Initialization.}

As discussed above, \method is initialized from DINOv3.
We examined this choice and found that, without DINO initialization (\ie, from scratch), the model can still converge to similar performance, but requires 4--8$\times$ more training iterations.
DINOv3 substantially eases optimization and improves training efficiency.

\paragraph{Invalid Area Prediction.}

We experimented with an additional branch for predicting invalid regions, as in~\cite{wang24moge:, keetha26mapanything:}, with the hope that it would help the model ignore regions such as sky and prevent sky pixels from sticking to foreground objects in the point cloud.
Although the model predicted the invalid masks accurately, sky pixels still appeared in the foreground of the depth estimates, likely due to the lack of supervisory signals in these regions.
Therefore, in line with the idea of using as few dense heads as possible, we did not include this predictor in our final model.

\paragraph{Register Attention Only.}

We also tested using register attention only, without any full global attention.
In this setting, image tokens in one frame can never attend to image tokens in other frames.
All the inter-frame information exchange relies on registers instead.
Although this design reduces FLOPs to 6\% of the original model, performance drops to the level of the original VGGT\@.
This trade-off may be worth exploring for on-device applications.

\paragraph{Auxiliary Inputs.}

Theoretically, incorporating auxiliary inputs, such as temporal order, camera parameters, depth maps, or scale factors, can further enhance performance.
However, we empirically observe that introducing these priors during pretraining, even when applied randomly or masked across training iterations, is often detrimental.
Conversely, our preliminary experiments indicate that providing conditional auxiliary inputs exclusively during the fine-tuning phase is highly effective, improving task-specific performance without compromising the integrity of the learned representations.
While a comprehensive exploration is beyond the scope of this paper, we believe this represents a promising direction for future research.

\paragraph{Synthetic and Real Data Mixture.}

Consistent with the observations in VGGT~\cite{wang25vggt}, we find that synthetic and real data play complementary roles during training.
At a high level, although this distinction is not exact, synthetic data contributes more directly to accuracy, while real data improves generalization.
The main role of real data is to adapt the model to real-world appearance and expose it to more diverse camera trajectories.
In practice, we recommend sampling roughly 80\% synthetic data and 20\% real data in each epoch.
If sufficiently clean synthetic annotations are available, \eg after addressing the issues discussed in~\cref{sec:data_quality}, increasing the synthetic data ratio to around 90\% may be even better.

\paragraph{How to fine-tune VGGT/\method.}

When fine-tuning VGGT/\method for reconstruction tasks in a new data domain, we recommend two practical choices.
First, it is best to use a full learning-rate schedule, \ie, linear warm-up followed by cosine decay.
The peak learning rate should remain small, and the number of iterations need not be large, but completing the full schedule is typically preferable to using a constant learning rate.
Second, we found that the aleatoric uncertainty (confidence) loss can be unstable during fine-tuning.
We therefore recommend initially removing this loss, particularly when the fine-tuning dataset is very small.
If the downstream task requires confidence estimates, the uncertainty head can be fine-tuned separately.
When fine-tuning VGGT/\method for tasks besides reconstruction, we suggest increasing the warm-up ratio, \eg from 5\% to 10--15\%, and use a full scheduling cycle.

\paragraph{Self-Supervised Training.} 
In principle, self-supervised training is a powerful tool to scale training data.
So far, we have found it useful for improving model generalization, especially for out-of-distribution data, however, it has had little impact on most benchmarks.
It is non-trivial to do better.
Concurrent work has made a similar observation~\cite{huang2026self}. 
We spent considerable time exploring other self-supervised protocols such as new view synthesis, including variants similar to RayZer~\cite{jiang25rayzer:} and E-RayZer~\cite{zhao26e-rayzer:}, generating tokens instead of pixels, NeRF representations~\cite{barron21mip-nerf:} or Gaussian Splats~\cite{kerbl233d-gaussian}. 
We also tried masking image tokens, distinguishing objects across frames, incorporating temporal order, and related variants. 
Only the student-teacher approach helped, in our implementation. 
For example, we found methods like E-RayZer, which may work well with static scenes, struggle with dynamic ones.
This was counterintuitive to us, because much of self-supervised training in 2D relies on some form of \textit{invariance} (\eg, image augmentation in DINO), and reconstruction should naturally encode such invariance.  
We expected self-supervised reconstruction to work from scratch, whereas our successful approach still requires a pretrained model. 
Therefore, although the teacher-student self-supervised training in~\cref{sec:self-supervised-training} is not necessary to achieve our benchmark results, we still share it with the community, even though it makes our training pipeline look more complicated. 
Self-supervised reconstruction remains an open problem for the community. 
It may ultimately need to be integrated into a unified or omni-style model, as we discuss later.

\paragraph{Dense Prediction Heads.}

In principle, strong latent representations should allow dense prediction tasks to be decoded with MLPs alone, without relying on convolutional layers, as evidenced by recent results such as JiT~\cite{li26back} or LagerNVS~\cite{szymanowicz26lagernvs}.
We explored this idea for a time because it aligned with our goal of keeping the framework as simple and scalable as possible.
However, we found that MLP-only heads consistently produced visible patch/block artifacts in the predicted depth maps, although they were often better than convolution-based alternatives in quantitative metrics, were much faster and more memory-efficient, and provided more stable gradients during training.
These artifacts are especially noticeable to humans, as they introduce geometric discontinuities that would otherwise be smooth.
We tried several remedies, including mipmap-style supervision~\cite{barron21mip-nerf:} (introducing local depth variance), probabilistic modeling (estimating depth with a probabilistic mixture of various independent channels),  and related variants, but they did not reliably remove the artifacts. 
Interestingly, the problem was much more prevalent in outdoor scenes than in indoor scenes, especially when the scene contained distant objects.
We suspect that this behavior is related to the numerical distribution of the prediction target.
For image generation tasks as studied by JiT, the output lies in a well-bounded numerical space.
In contrast, the range of geometric quantities, such as depth, is effectively unbounded.
This has motivated our trade-off, where we maintain a small number of convolutional layers followed by MLP layers.
The shallow convolutions operate at low feature resolutions, such as $16\times16$ or $32\times32$, and introduce little computational overhead, while accounting for most of the observed improvement in spatial smoothness. 
Notably, we used the shallow layers of DPT to keep the architecture comparable to prior methods, but any feature-pyramid-style convolution architecture can address the patch-artifact problem. 
Although we use the trade-off for \method, we still believe that MLP-only dense decoding heads are a promising and important direction for future exploration.

\section{Discussion}%
\label{sec:option}

We now discuss some decisions in designing \method, which are in part motivated by how we think feed-forward models can be most beneficial to the community.

\paragraph{Prioritizing Simplicity.}

We identified several architectural modifications that can further improve performance.
For example, while our model predicts the camera parameters in a single pass, better results can be obtained by using the iterative refinement design of the original VGGT~\cite{wang25vggt}.
The depth output can also be improved by injecting raw RGB values into the dense prediction head.
Using these and other techniques, we observed further improvements of 4\%--6\% in AUC@$3^\circ$ and about 2\% in \(\delta_{1.25}\) across multiple datasets, and we expect that further task-specific modifications can yield even more gains.
Even so, we deliberately decided to prioritize the overall simplicity of the model and the quality of the representation extracted by the feature backbone (aggregator). 
This choice is motivated by our observation that, once the backbone is well-trained, training a new or improved prediction head typically requires only 5--10K iterations.
Additionally, maintaining the simplicity of the architecture yields a cleaner base model, which we expect will be easier for the community to build upon.

\paragraph{Benefits of Feed-forward Reconstruction.}

Compared to traditional reconstruction pipelines, feed-forward reconstruction has three important advantages:
(a) efficiency,
(b) robustness, and
(c) representational power.
First, it is substantially faster, from tens to hundreds of times in some settings, and the gap can become even larger by streaming or by using multiple GPUs in parallel.
Second, it is much more robust than methods that rely heavily on explicit geometry optimization, like bundle adjustment.
For example, it can handle cases with little or no parallax, which are difficult to handle with triangulation-based methods.
Feed-forward reconstruction also readily extends to processing dynamic scenes, as shown in our experiments. 
Third, and perhaps most importantly, feed-forward reconstruction can extract versatile geometry-aware representations.
3D vision lacks powerful off-the-shelf general-purpose feature extractors comparable to the older VGG16~\cite{simonyan15very} or ResNet~\cite{he16deep} for 2D vision, let alone to modern 2D vision or vision-language foundation models.
As we show with several examples in this paper, feed-forward reconstruction is a promising proxy task for learning such a representation for 3D tasks.

Optimization-based reconstruction methods like COLMAP~\cite{schonberger16structure-from-motion} still have their strengths.
In well-conditioned settings, bundle adjustment can estimate the camera parameters to extremely high precision, reducing angular errors to a few hundredths of a degree.
Such accuracy remains valuable for applications like novel view synthesis with NeRFs~\cite{mildenhall20nerf:} or Gaussian splatting~\cite{kerbl233d-gaussian}.
Feed-forward reconstruction is not in conflict with optimization, instead, it can serve as a strong initialization for subsequent optimization procedures such as bundle adjustment.
Even so, given the rapid progress of feed-forward reconstruction in the short time since VGGT was introduced, we remain optimistic about future accuracy improvements and the applicability of this paradigm.

\paragraph{The Role of 3D/4D in the Era of Large Models.}

Spatial understanding is an area of increasing interest in vision-language, vision-language-action, vision-action, and world models~\cite{
chen24spatialvlm:,cheng24spatialrgpt:,zitkovich23rt-2:,cai26depthlm:,kim25openvla:,
team25gemini,hu23gaia-1:,li26causal,parker-holder25genie,nvidia25cosmos}. 
This is a natural consequence of developing embodied agents that operate in the physical world, which is three-dimensional.
In the short term, reconstruction models can be used as external tools to obtain explicit 3D information, such as depth and camera parameters.  
They can also provide `structured' tokens that encode spatial information implicitly, as discussed in this paper. 
In the long term, reconstruction might become a citizen in future unified models (`omni-model'), \ie, in large models pre-trained using multiple modalities, such as language and vision, and targeting multiple capabilities, such as understanding and generation~\cite{wang24emu3:,zhou25transfusion:,tong26beyond,team26longcat-next:,wang26ernie,team24chameleon:,wu25janus:,deng25emerging}. 
This could be achieved by adding reconstruction-oriented data to model training without significant changes to the architecture or training framework. 
For example, camera parameters can be predicted autoregressively as text, while pixel-wise quantities such as depth can be formulated as image generation, as recently explored in~\cite{yang26context}.

\paragraph{The next generation of reconstruction models, and perhaps more broadly perception systems, may be built on unified models.}
First, the biggest gain is likely to come from data, which is currently the main bottleneck for perception tasks like reconstruction.
Large-scale text and video corpora contain rich implicit descriptions of the physical world.
When geometry tasks are trained jointly with these modalities, they can tap into this massive, broader source of supervision.
Second, most perception problems are severely under-constrained when viewed in isolation.
A unified model naturally enforces cross-task consistency, allowing ambiguities in one domain (\eg, textureless regions in depth estimation) to be resolved by priors from another (\eg, semantic context).
Third, there is a paradigm-level reason to move in this direction: recent observations may indicate generative vision models scale more easily than perception-only vision models, and it seems generative models can transfer to perception to some extent~\cite{gabeur26image}.
We expect stronger, more robust perception models to emerge from multi-modal, multi-task training, rather than from reconstruction or perception objectives in isolation.

\section{Conclusion}%
\label{sec:conclusion}

We presented \method, a feed-forward reconstruction model that achieves strong results across static and dynamic benchmarks.
We improved the original VGGT in terms of architecture, data, and training by introducing register attention, using a single dense prediction head with multi-task losses, a large-scale annotation pipeline that handles dynamic content, and a self-supervised training protocol that leverages vast amounts of unlabeled videos.
These ingredients allowed us to train our model at an unprecedented scale.
Empirically, we found that \method scales predictably with model capacity and data size.
Beyond geometry, we found that the learned registers carry useful global information, improving VLA models and supporting alignment with language.
We hope \method will be a useful foundation for the community to build on.

\paragraph{Acknowledgments.}
Jianyuan Wang was supported by Facebook AI Research. 
Christian Rupprecht was partially supported by ERC starting grant `Volute' (No. 101222037).
Many people helped bring this work together. 
Please see our project page for full acknowledgments.

{
\small
\bibliographystyle{ieeenat_fullname}
\bibliography{vedaldi_general,vedaldi_specific,main}
}

\clearpage
\maketitlesupplementary
\appendix

In this Supplement, we provide additional implementation details in \cref{sec:impl}, a discussion of common data issues in \cref{sec:data_quality}, and a discussion of limitations in \cref{sec:limitation}.

\section{Additional Details}%
\label{sec:impl}

In this section, we detail the training setup and architecture (\cref{sec:impl-training}), the construction of positive and negative pairs (\cref{sec:impl-pairs}), the VLM prompt (\cref{sec:impl-vlm-prompt}), the annotation filtering criteria (\cref{sec:impl-filtering}), and the language alignment procedure (\cref{sec:impl-language-alignment}).

\subsection{Training and Architecture}%
\label{sec:impl-training}

To train our model, we set the loss weights in \cref{eq:training-loss} to $\lambda_{\text{cam}} = 5.0$, $\lambda_{\text{depth}} = 1.0$, $\lambda_{\text{point}} = 0.5$, and $\lambda_{\text{match}} = 0.1$.
For each scene, following~\cite{wang24dust3r:, wang25vggt}, we normalize the ground truth to a unit space.
Concretely, we first transform all quantities into the first camera's coordinate frame and compute the average distance of all 3D points to the origin; we then scale the depth maps and translation vectors by this value.
As in~\cite{wang25vggt}, this normalization is applied only to the ground truth and not to the predictions.
To stabilize training, we apply gradient-norm clipping with a threshold of $1.0$ and use QKNorm inside the attention layers.
At each iteration, we randomly sample the number of input frames from the range $[1, 24]$ and correspondingly set the batch size to saturate GPU memory.
Each frame is independently masked by a rectangle whose height and width are uniformly sampled from $[32, 128]$ pixels, with a probability of $0.05$.
Pixels inside the masked region are set to black, and the corresponding depth values are marked as invalid.
For color jittering, we use PyTorch's implementation with brightness $0.5$, contrast $0.5$, saturation $0.5$, and hue $0.1$.
During self-supervised training, the weights of the teacher model are updated using an exponential moving average with a decay of $0.999$.
Our camera head follows the implementation in~\cite{wang25vggt}, using a ReLU activation for the focal length and no activation for the quaternion and translation parameters.
For datasets with temporally ordered sequences (\eg, videos), we sample frames from a local temporal window rather than based on covisibility.
This can produce particularly challenging reconstruction subsets in which images are semantically related but share little or no visual overlap.
We found that these samples improve the model's generalization.

\subsection{Constructing Positive and Negative Pairs}%
\label{sec:impl-pairs}

We expand on the matching loss introduced in the main paper (\cref{sec:training-losses}) by detailing how we construct the positive and negative token pairs.

Given per-pixel 3D points obtained by unprojecting the depth maps, we first sample valid pixels in the query frame and assign each sampled pixel to a query patch.
Using the known camera intrinsics and extrinsics, we then project the corresponding 3D points into all other frames and retain only those projections that (i) fall inside the image and (ii) are depth-consistent with the target-frame depth maps (within a small relative tolerance of $1\%$ and excluding a narrow image boundary of 4 pixels).

For each target frame, we count how many 3D points from each query patch land in each target patch.
This yields, for every query patch, a soft correspondence map over target patches in the form of projection overlap ratios.
We select query patches that have sufficient valid projections across views and randomly sample up to a fixed number of them, with sampling probabilities proportional to their total number of correspondences.
For each selected query patch, all target patches with $>$10\% overlap form positive token pairs.

Because some sequences are dynamic, failing the positive-pair criteria does not automatically define a valid negative.
Instead, we construct negative pairs by randomly sampling patches as query and target candidates, and then enforcing two constraints:
(i) a geometric constraint, requiring the candidate target patch center to lie sufficiently far from the epipolar line induced by the query patch (\ie, to have a large epipolar/Sampson distance), and
(ii) an appearance constraint, requiring a sufficiently large \(\ell_2\) distance between the mean RGB values of the two patches.
Patches that satisfy both constraints are treated as negatives.
We then subsample these negatives to obtain a balanced set of positive and negative patch pairs for supervising the matching loss on the last-layer tokens.

\subsection{VLM Prompt}%
\label{sec:impl-vlm-prompt}

We use a VLM to discard videos not suitable for multi-view geometry.
We prompt the VLM as follows:

{
  \vspace{1em}
  \color{linegreen}
  \hrule
  \vspace{1pt}
  \hrule
  \vspace{0.5em}
}

\begingroup

\color{promptcolor} \ttfamily \footnotesize \raggedright
\noindent
"You are an expert in computer vision, photogrammetry, and Multi-View Geometry (MVG)."\\
"Analyze the video clip to determine if it is suitable for high-quality 3D reconstruction."\\
"You must categorize the video and extract scene metadata based on the strict criteria below."

\vspace{0.5em}
\noindent {STEP 1: CHECK FOR HARD REJECTION FACTORS} \\
``Assess the following. If ANY are present, the Classification is {`REJECT\_HARD'}'':
\begin{enumerate}[noitemsep, topsep=0pt, leftmargin=*]
    \item Discontinuities \& Editing: Does the video contain cuts, dissolves, wipes, fades, or montage editing? (Must be a single continuous shot).
    \item Non-Physical/2D Content: Is this a screen recording, a slideshow of static images, animation, or a cartoon? (Must be real-world footage).
    \item Extreme Visual Failure: Severe motion blur (edges lost), severe rolling shutter (wobble/jello effect), or corruption/glitching.
    \item Major Obstructions: Are there heavy overlays (large text/UI), watermarks, or physical obstructions like a finger covering the lens?
    \item Non-Pinhole Projections: Is the footage 360$^{\circ}$ equirectangular or heavily distorted fisheye without calibration?
\end{enumerate}

\vspace{0.5em}
\noindent {STEP 2: CHECK FOR GEOMETRIC \& TEXTURAL SUITABILITY} \\
"If the video passes Step 1, assess for reconstruction quality. If ANY are present, the Classification is {`REJECT\_SOFT'}:"
\begin{enumerate}[noitemsep, topsep=0pt, leftmargin=*]
    \item Insufficient Parallax: Is the camera stationary? Does it only rotate (pan/tilt) or zoom without physical translation? (Translation is required for depth).
    \item Texture Issues: Does the scene lack non-repetitive texture? (e.g., blank white walls, clear blue sky only, dark shadows, or highly repetitive patterns like grid tiles).
    \item Specularities \& Transparencies: Are the dominant features reflective (mirrors, water, glass) or transparent? (These create `ghost' geometry).
    \item Focus \& DOF Issues: Is there severe focus hunting (pulsing), or is the depth-of-field so shallow that the background is entirely blurred (bokeh)?
    \item Dynamic Dominance: Is the frame dominated ($>$90$\%$) by a moving subject (e.g., a close-up talking head) while the static background is visible but minimal?
\end{enumerate}

\vspace{0.5em}
\noindent {STEP 3: EXTRACT METADATA}\\
"Determine the scene dynamics:"
\begin{enumerate}[noitemsep, topsep=0pt, leftmargin=*]
    \item Dynamic: Moving objects are present (people walking, cars moving, wind blowing trees heavily).
    \item Static: The scene is rigid; only the camera is moving.
\end{enumerate}

\vspace{0.5em}
\noindent {OUTPUT INSTRUCTIONS:} \\
"Return a JSON object strictly following this schema. Do not output markdown formatting or explanations outside the JSON."
\begin{verbatim}
{
"classification": 
"ACCEPT" | "REJECT_HARD" | "REJECT_SOFT",
"reason": "Brief descriptions of the primary 
flaw or 'Good candidate'",
"scene_dynamics": "Static" | "Dynamic"
}
\end{verbatim}

\noindent "Label Definitions:" \\
{REJECT\_HARD}: Unusable due to editing, synthetic content, severe artifacts, or obstructions. \\
{REJECT\_SOFT}: Technically usable but risks failure due to rotation-only, reflections, lack of texture, or focus issues. \\
{ACCEPT}: High-quality candidate. Single shot, clear translation/parallax, good texture, rigid geometry.
\endgroup

{
  \vspace{0.5em}
  \color{linegreen}
  \hrule
  \vspace{1pt}
  \hrule
  \vspace{1em}
}

\subsection{Annotation Filtering Criteria}%
\label{sec:impl-filtering}

To train the ensemble classifier (XGBoost, Random Forest, and CatBoost) described in \cref{sec:annotation-pipeline}, we extract several geometric features from each reconstruction.
These features are designed to detect common failure modes such as collinear motion degeneracies, inconsistent scale, and noisy camera trajectories.
Below, we provide definitions for the most important features implemented in our pipeline.

\paragraph{Trajectory Smoothness.}

We quantify the smoothness of the estimated camera trajectory using the acceleration of the translation vectors $\mathbf{t}$.
The smoothness score $S_\text{trans}$ is calculated as the mean squared magnitude of the acceleration:
\begin{equation}
    S_\text{trans} = \frac{1}{N-2} \sum_{i=1}^{N-2} \| \mathbf{t}_{i+1} - 2\mathbf{t}_i + \mathbf{t}_{i-1} \|^2
\end{equation}
We compute a similar metric $S_\text{rot}$ for rotations using the second order difference of the rotation vectors.
High values in $S_\text{trans}$ or $S_\text{rot}$ indicate jittery trajectories, often associated with poor SfM convergence.

\paragraph{Parallax Angle Analysis.}

To ensure sufficient baseline for multi-view stereo, we analyze the parallax angles of the reconstructed sparse point cloud.
For a subset of sparse points $\mathcal{P}$, we identify the set of cameras $\mathcal{C}_p$ visible to point $p \in \mathcal{P}$.
We compute the maximum angle subtended by any pair of cameras $c_j, c_k \in \mathcal{C}_p$ at point $p$.
The final feature is the {median} of these maximum angles across all sampled points.
Low median parallax indicates degenerate rotation-only motion or extreme distance.

\paragraph{Point Cloud PCA Shape (Linearity \& Planarity).}

To detect geometric degeneracies such as ``fly-by'' straight-line reconstructions (which result in cylindrical ambiguity), we perform Principal Component Analysis (PCA) on the normalized point cloud coordinates.
Let $v_1 \ge v_2 \ge v_3$ be the eigenvalues of the point cloud covariance matrix.
We define \emph{Linearity} as $(v_1 - v_2) / v_1$, \emph{Planarity} as $(v_2 - v_3) / v_1$, and \emph{Scattering} as $v_3 / v_1$.
Reconstructions with excessively high linearity are flagged as linear degeneracies and are typically discarded by the classifier.

\paragraph{Depth Map Completeness.}

This metric evaluates the density of the dense reconstruction stage.
For each frame, we calculate the percentage of pixels containing valid, finite depth values relative to the total image resolution.
The final feature is the average completeness across all frames.
Extremely low values typically indicate failure in the Multi-View Stereo stage, often caused by textureless surfaces, high specularities, or bad cameras.

\paragraph{Point Cloud Noise Level.}

We estimate the signal-to-noise ratio using statistical outlier detection.
For every point, we compute the average distance to its nearest neighbors.
Points are classified as noise if this distance exceeds the global average by more than two standard deviations.
The feature is the percentage of points classified as noise, which allows us to detect reconstructions plagued by floating artifacts and outliers.

Note that this pipeline can provide depth values only for rigid pixels, so the metrics above assume that dynamic pixels are already masked out.
We assume the model can learn to estimate depth for dynamic pixels from synthetic data.

\subsection{Language Alignment}%
\label{sec:impl-language-alignment}

In \cref{sec:applications}, we use a VLM to produce a sequence-level language embedding for each training sequence.
Here, we provide the exact VLM prompt and representative generated descriptions.
The hidden states of the generated text tokens are mean-pooled and projected to form the language embedding.
The prompt and image tokens are excluded from this pooling.
We use at most \(128\) newly generated tokens and keep the VLM fixed.
The \method side is optimized during alignment fine-tuning, as described in \cref{sec:applications}.
The images are provided as an unordered collection, both to the VLM and to our model.

{
  \vspace{1em}
  \color{linegreen}
  \hrule
  \vspace{1pt}
  \hrule
  \vspace{0.5em}
}

\begingroup

\color{promptcolor} \ttfamily \footnotesize \raggedright
\noindent {\normalfont\bfseries Language Alignment Prompt}

\vspace{0.25em}
\noindent
These images show the same scene from multiple viewpoints.\\
Describe the full scene as one coherent scene, not as separate images.\par\vspace{0.5em}\noindent
Use the following format:\\
Scene: \textless{}scene category or environment\textgreater{}.\\
Content: \textless{}main objects and coarse spatial arrangement\textgreater{}.\\
Appearance: \textless{}stable visual attributes such as material, color, or style\textgreater{}.\par\vspace{0.5em}\noindent
Write one short sentence for each field.\\
Mention information that is consistent across multiple views, even if it is not visible in every view.\\
Focus on shared scene content, major objects, coarse layout, and stable appearance.\\
Mention dynamic objects only if they are prominent and identifiable in multiple views.\\
Do not describe motion, actions, or frame-specific changes.\\
Do not describe each image separately.\\
Do not mention camera motion, viewpoint order, image quality, blur, exposure, lighting artifacts, or uncertainty.\\
Keep the description concise, factual, and semantically dense.

\vspace{0.7em}
{\color{linegreen}\hrule height 0.6pt}
\vspace{0.6em}
\noindent {\normalfont\bfseries Representative Descriptions}
\vspace{-0.6em}
\newcommand{\exampleDashedSep}{\par\vspace{0.4em}\noindent{\color{linegreen}\leaders\hbox{\rule{0.8em}{0.4pt}\hspace{0.35em}}\hfill\kern0pt}\par\vspace{0.4em}}

\exampleDashedSep
\noindent Scene: A dimly lit room with a red hue, featuring a door, a window, and a wall with a patterned texture.\\
Content: The room contains a closed door on the left, a window on the right, and a wall with a textured pattern.\\
Appearance: The walls have a floral or damask pattern, the floor is dark, and the overall lighting is red.

\exampleDashedSep
\noindent Scene: Snowy forest landscape with a winding road.\\
Content: A snow-covered forest with tall pine trees, a winding road, and a steep slope.\\
Appearance: The scene has a high-contrast look, with dark clouds above and bright white snow covering the ground and trees.

\exampleDashedSep
\noindent Scene: Outdoor public area.\\
Content: A dark green plastic picnic table with a hexagonal top and black legs, accompanied by matching benches, situated on a paved surface.\\
Appearance: The table and benches are made of uniform dark green plastic with a matte finish, and the ground is composed of light gray square tiles.

\exampleDashedSep
\noindent Scene: Urban plaza with modern architecture.\\
Content: A central fountain with a large circular sculpture, surrounded by paved walkways and buildings.\\
Appearance: The fountain has a metallic, silver-colored structure with a dark base, and the surrounding area features light gray stone tiles and modern buildings with glass and concrete facades.

\endgroup

{
  \vspace{0.5em}
  \color{linegreen}
  \hrule
  \vspace{1pt}
  \hrule
  \vspace{1em}
}

\section{Data Quality}%
\label{sec:data_quality}

One of the most important ingredients for training large `foundation' models is data.
Having access to a large amount of data is, however, insufficient; in fact, the \emph{quality} of the data also has a strong effect on the model's behavior.
For 3D reconstruction, we found that noisy data introduces particular failure modes in both self-supervised and supervised training.

For self-supervised training, we observe that unreasonably difficult videos, such as the ones containing discontinuous shot transitions as commonly seen in television footage, induce sharp loss spikes and may cause gradients to explode, resulting in a degradation of the model's performance.
We therefore restrict self-supervised training to videos that pass the VLM pre-filtering check of \cref{sec:annotation-pipeline}, which reliably removes such cases.

In the supervised stage, the effects of noisy data are more subtle but equally important.
We find that different types of errors in the annotations translate into specific failure modes at inference time.
What is worse, these failure modes do not affect most of the images in standard benchmarks and may thus not be detected in the quantitative results.
Instead, they emerge only when the model is tested on a new sample that resembles an incorrectly labeled sample seen during training.
This behavior indicates that, while the model does learn the general principles of 3D reconstruction, it can nevertheless memorize idiosyncratic noise as well.
Next, we discuss the most representative cases we have encountered.

\begin{figure}
\centering
\includegraphics[width=0.95\linewidth]{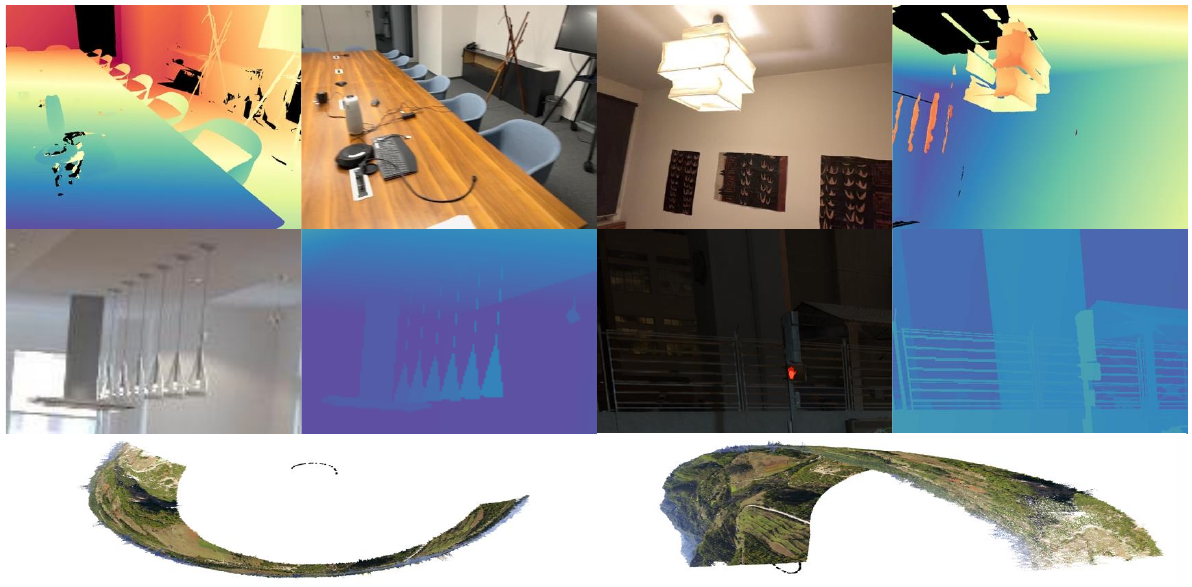}
\caption{\textbf{Common Data Issues.}
Top: examples from ScanNet++.
Middle: examples from synthetic datasets such as Hypersim.
Bottom: the doming effect.}%
\label{fig:data_quality}
\vspace{-8pt}
\end{figure}

\paragraph{Sensors.}

Some datasets annotate real-world scenes using sensors such as LiDAR or depth cameras.
A common failure mode here is foreground-background leakage.
For example, as shown in the top row of~\cref{fig:data_quality} with an example from ScanNet++, the depth of the chair's back corresponds to the background floor or wall, likely due to misalignment or holes in the sensor capture.
We also observe spurious depth artifacts around the hanging light fixture, where the rendered depth contains fragmented foreground structures that do not correspond to the visible lamp geometry.
Such artifacts are likely caused by unreliable sensor capture or reconstruction around over-exposed and partially translucent objects.
We therefore exclude such datasets in the later phase of training.

\paragraph{Thin Structure.}

Synthetic data usually provides more accurate geometric annotations, but thin structures can be inaccurate in some synthetic datasets too.
Objects such as fence bars often occupy only a few pixels and lie on sharp depth discontinuities.
Although these structures are clearly visible in the RGB image, their rendered depth can be incomplete, overly smoothed, or misaligned.
As a result, the label may assign a thin structure to the depth of the wall, window, floor, or distant background behind it, rather than to the actual object surface, as shown in the middle row of~\cref{fig:data_quality}.
At inference time, the model may ignore thin objects, recover only their thicker or lower parts, or produce depth maps in which thin structures are washed out by nearby large surfaces.
This is a common problem in existing models, and we hypothesize that it can be mitigated with higher-resolution inputs or better synthetic data.

\paragraph{Fake Background.}

Another common issue in synthetic data is fake background.
For example, in Kubric, PointOdyssey, BEDLAM, the background depth may correspond to proxy dome or floor geometry used for HDRI rendering, rather than the true 3D structure suggested by the appearance of the background.
Although this is reasonable for rendering, it provides depth values that are inconsistent with the scene semantics.
For datasets with a clear boundary between the real foreground and the synthetic background, such as BEDLAM, we filter out the background during training by thresholding the maximum valid foreground depth.
We exclude the remaining from training.

\paragraph{Doming Effect.}

Methods using bundle adjustment, such as COLMAP~\cite{schonberger16structure-from-motion}, MegaSaM~\cite{li25megasam:}, and ViPE~\cite{huang25vipe:}, can be used to generate pseudo ground truth.
However, care is needed because they may produce degenerate solutions such as the doming effect, as shown in the bottom row of~\cref{fig:data_quality}.
This usually arises from weak global constraints in the image collection, \eg, near-parallel viewing directions, insufficient loop closure, small triangulation angles, or inaccurate camera self-calibration, especially radial-distortion errors, which are common in Internet videos.
Under these conditions, bundle adjustment can still achieve a low reprojection error while bending the recovered cameras and scene geometry into a globally inconsistent curved shape.
Such reconstructions may appear locally plausible but provide incorrect large-scale geometry, making them harmful as supervision.
They can be filtered out by supervised geometric filtering, as discussed in~\cref{sec:data}, or by comparing the annotated depths with predictions from existing monocular depth models or feed-forward reconstruction models.

\paragraph{Humans in Walls.}

We observe a recurring failure case in near-static street-view videos with pedestrians moving through the scene: almost all existing feed-forward reconstruction models often absorb humans into the static background, estimating them as part of nearby walls or buildings.
This failure appears to be related to ambiguous boundary pixels in existing training datasets.
For example, the most widely used multiview dataset Megadepth was annotated with COLMAP on phototourism images that often contain people.
At human boundaries, the patch match stereo algorithm may assign some pixels to the surrounding static architecture, causing supervision to treat parts of people as background.
Excluding MegaDepth or re-estimating its depths can substantially reduce this artifact.

\paragraph{Data Ambiguities.}

Inconsistencies in annotations across different datasets can lead to confusing model behaviors.
For instance, in synthetic datasets such as Aria Synthetic Environments and HyperSim, outdoor scenes are often rendered as 2D textures applied to windows, meaning the GT depth corresponds to the window surface itself.
Conversely, in other datasets, depth values correctly represent the actual physical objects visible through windows.
This inherent GT mismatch across the training corpus confuses the model, which occasionally leads to strange predictive behaviors.

Several recent datasets have introduced alternative pipelines for annotating Internet videos, including SceneScribe-1M~\cite{wang26scenescribe-1m:}, PointWorld~\cite{huang26pointworld:}, Sekai~\cite{li25sekai:}, SpatialVID~\cite{wang25spatialvid:}, and OmniWorld~\cite{zhou26omniworld:}, which may also be worth exploring.

\section{Limitations}%
\label{sec:limitation}

\paragraph{Failure Cases.}

Despite the generally strong performance of \method, we observe specific scenarios in which the model struggles. 
For example, the model's performance drops significantly in the presence of strong motion blur. 
Meanwhile, reconstruction quality often degrades if the field of view changes abruptly (e.g., shifting from $10^\circ$ to $160^\circ$ in a few seconds) or the camera is highly distorted. 
Additionally, because the model was exposed to some noisy data (\eg, ScanNet++) during the early stage of training, its predictions are sometimes unstable in the cases like office scenes with many monitors. 
These limitations are primarily attributable to the distribution of our training data, and we hope to alleviate them in future work by incorporating more diverse and challenging sequences.

\paragraph{Masked Sensitive Content.}

Due to privacy and licensing constraints, some portions of the training data, such as human faces and trademarks, are masked or blurred.
As a result, the model may rarely produce unexpected artifacts or unstable predictions in these regions. For example, we observed that the predicted depth for a person wearing black clothing can occasionally become less smooth.
These artifacts can lead to visually distorted or qualitatively unappealing outputs.

\end{document}